\documentclass{article}

    \usepackage[preprint]{neurips_2025}

\usepackage[utf8]{inputenc} % allow utf-8 input
\usepackage[T1]{fontenc}    % use 8-bit T1 fonts
\usepackage{hyperref}       % hyperlinks
\usepackage{url}            % simple URL typesetting
\usepackage{booktabs}       % professional-quality tables
\usepackage{amsfonts}       % blackboard math symbols
\usepackage{nicefrac}       % compact symbols for 1/2, etc.
\usepackage{microtype}      % microtypography
\usepackage{xcolor}         % colors

\usepackage{bbm}
\usepackage{amsmath, amssymb}
\usepackage{algorithm}
\usepackage{algpseudocode}
\usepackage{mathtools}
\usepackage{bm}
\usepackage{subcaption}
\usepackage{multirow}
\usepackage{wrapfig}
\usepackage{multirow}

\title{Gradient Flow Matching for Learning Update Dynamics in Neural Network Training}

\author{%
  Xiao Shou \\
  Department of Computer Science\\
  Baylor University\\
  Waco, TX 76706 \\
  \texttt{xiao\_shou@baylor.edu} \\
  \And
  Yanna Ding \\
  Department of Computer Science\\
  RPI \\
  Troy, NY 12180 \\
  \texttt{dingy6@rpi.edu} 
  \AND
  Jianxi Gao \\
Department of Computer Science\\
  RPI \\
  Troy, NY 12180 \\
  \texttt{gaoj8@rpi.edu} \\
}

\begin{document}

\maketitle

% \begin{abstract}
% Training deep neural networks remains computationally intensive due to the iterative nature of gradient-based optimization. We propose \textbf{Gradient Flow Matching (GFM)}, a generative framework that models the evolution of neural network weights as a continuous-time dynamical system. By leveraging conditional flow matching, GFM learns optimizer-aware vector fields that capture the implicit update rules underlying SGD and adaptive optimizers like Adam and RMSprop. Unlike black-box sequence models, GFM incorporates the structure of gradient-based updates into the learning objective, enabling accurate forecasting of final weights from partial trajectories. Empirically, GFM achieves forecasting accuracy that is competitive with Transformer-based models and significantly outperforms LSTM and other classical baselines. Furthermore, GFM generalizes across neural architectures and initializations, providing a unified framework for studying optimization dynamics and accelerating convergence prediction.
% \end{abstract}
\begin{abstract}
Training deep neural networks remains computationally intensive due to the iterative nature of gradient-based optimization. We propose \textbf{Gradient Flow Matching (GFM)}, a continuous-time modeling framework that treats neural network training as a dynamical system governed by learned optimizer-aware vector fields. By leveraging conditional flow matching, GFM captures the underlying update rules of optimizers such as SGD, Adam, and RMSprop, enabling smooth extrapolation of weight trajectories toward convergence. Unlike black-box sequence models, GFM incorporates structural knowledge of gradient-based updates into the learning objective, facilitating accurate forecasting of final weights from partial training sequences. Empirically, GFM achieves forecasting accuracy that is competitive with Transformer-based models and significantly outperforms LSTM and other classical baselines. Furthermore, GFM generalizes across neural architectures and initializations, providing a unified framework for studying optimization dynamics and accelerating convergence prediction.
\end{abstract}

\section{Introduction}

Deep neural networks (DNNs) have achieved remarkable success in tasks such as image classification~\citep{lecun1998gradient} and natural language processing~\citep{vaswani2017attention}, driven by increasingly deep and complex architectures. As models scale to hundreds of layers and billions of parameters, training them requires vast computational resources and prolonged optimization cycles. The resulting workloads scale non-linearly with model size, creating serious bottlenecks in time and resource efficiency.

While inference has benefited from acceleration techniques like model compression~\citep{tan2019efficientnet} and knowledge distillation~\citep{sanh2019distilbert}, improving training efficiency remains a more fundamental challenge. This is largely because gradient-based optimization is inherently iterative: algorithms like SGD and Adam~\citep{kingma2014adam} update weights step-by-step, requiring thousands of iterations to reach convergence. Increasing hardware capacity helps, but cannot fully eliminate the cost of these sequential updates. Addressing this bottleneck calls for methods that rethink how optimization trajectories are modeled and predicted.

Recent studies have revealed that weight evolution during training exhibits predictable patterns, opening opportunities for forecasting-based acceleration. For instance, the Weight Nowcaster Network (WNN)~\citep{jang2023learning} predicts near-future weights to skip small optimization steps, while LFD-2~\citep{shou2025less} performs long-horizon "farcasting" from partial weight sequences. However, these methods treat optimization as a generic sequence modeling problem, without explicitly modeling the underlying dynamics of gradient-based updates. Consequently, they overlook the structured relationship between gradients, learning rates, and weight trajectories that governs the optimization process itself.

In this work, we propose \emph{Gradient Flow Matching} (GFM), a continuous-time framework that models neural network training as a dynamical system governed by optimizer-aware vector fields. Leveraging conditional flow matching~\citep{tong2023conditional}, GFM reinterprets weight trajectories not as arbitrary sequences, but as realizations of structured flows conditioned on partial observations and optimization algorithms. This perspective enables more accurate and generalizable forecasting of final weights by explicitly incorporating the dynamics of gradient-based updates. Our main contributions are as follows:
\begin{itemize}
    \item \textbf{A flow-matching-compatible formulation of first-order dynamics.} While continuous-time views of gradient descent are well-studied, we tailor them specifically for conditional flow matching, enabling the learning of vector fields that are both optimizer-aware and trajectory-faithful.

    \item \textbf{A conditional flow matching loss for weight evolution.} We introduce a loss function that aligns learned vector fields with observed parameter updates, supporting both short-term extrapolation and long-horizon forecasting within a unified framework.

    \item \textbf{Extension to momentum and adaptive optimizers.} GFM accommodates common optimizers such as Adam~\citep{kingma2014adam} and RMSProp~\citep{shi2021rmsprop}, capturing their dynamics within a continuous-time approximation through the learned flow field.
\end{itemize}

By marrying weight trajectory forecasting with continuous-time modeling, GFM offers a unified and extensible lens for analyzing and predicting the evolution of neural network weights.

\section{Related Work}

% \paragraph{Weight Forecasting and Acceleration.}
% The idea of forecasting weights to accelerate training originated with Introspection~\citep{sinha2017introspection}, which used past training history to predict future weights. The Weight Nowcaster Network (WNN)~\citep{jang2023learning} extended this to periodic short-horizon forecasting that enabled skipping training steps. While effective, WNN and its variants primarily operate on small networks and forecast only a few steps ahead. To address long-range forecasting, Shou et al.~\citep{shou2025less} introduced "farcasting" with LFD-2, demonstrating that even a single-layer model can forecast weight trajectories over tens of epochs using just initial and final weights. Concurrently, \citep{knyazev2025nino} proposed a neuron-graph-based approach (NiNo) for conditional horizon forecasting, using GNNs to encode architectural structure. These models demonstrated that weight evolution is both compressible and predictable. Other acceleration strategies include meta-learned optimizers such as L2O~\citep{chen2022learning}, which train neural networks to produce optimizer updates. However, these require costly meta-training and incur per-step overhead. Lookahead~\citep{zhang2019lookahead} offers a simpler strategy by interleaving fast inner updates with slow averaging steps, leveraging anticipated future states without learning them directly.
\paragraph{Weight Forecasting and Acceleration.}
The idea of forecasting weights to accelerate training dates back to Introspection~\citep{sinha2017introspection}, which used past training history to predict future parameters. The Weight Nowcaster Network (WNN)~\citep{jang2023learning} extended this idea to short-horizon prediction, enabling periodic skipping of optimization steps. While effective on small-scale models, WNN and its variants typically forecast only a few steps ahead and treat each weight independently. To support long-range forecasting, Shou et al.~\citep{shou2025less} proposed "farcasting" with LFD-2, showing that even simple models can extrapolate full training trajectories from just the initial and final weights. Knyazev et al.~\citep{knyazev2025nino} \emph{concurrently} introduced a graph-based approach (NiNo) that incorporates architectural structure into the forecasting process. Other acceleration strategies include meta-learned optimizers such as L2O~\citep{chen2022learning}, which train neural networks to generate optimizer updates, albeit with significant meta-training cost. Simpler alternatives like Lookahead~\citep{zhang2019lookahead} reduce overhead by interleaving fast inner updates with slow outer updates, leveraging future trends without explicitly predicting them.

%As NiNo was developed concurrently, we do not include it as a baseline in our experiments.
% \paragraph{Flow-Based and Continuous-Time Models.}
% Continuous gradient flows offer an alternative view of optimization as an ODE-driven process. Neural ODEs~\citep{chen2018neural} have enabled continuous modeling of hidden states but are computationally intensive and not tailored to optimizer dynamics.
\paragraph{Flow-Based and Continuous-Time Models.}
Continuous gradient flows provide an alternative perspective on optimization by interpreting training as a trajectory governed by an ordinary differential equation (ODE). Neural ODEs~\citep{chen2018neural} enable continuous-time modeling of hidden states in deep networks, offering flexibility for \emph{irregularly} spaced observations. However, they can be computationally expensive due to the need for numerical integration at each time step, particularly in stiff or high-dimensional systems. Moreover, standard Neural ODEs are not specifically designed to capture the structure of gradient-based optimization dynamics. In practice, many forecasting methods discretize these trajectories using fixed steps, making them analogous to sequence models like LSTMs. Flow Matching (FM)~\citep{lipmanflow} and its conditional variant~\citep{tong2023conditional} offer a scalable, simulation-free way to learn continuous-time dynamics. %FM regresses a neural vector field to match known interpolation paths between distributions. Conditional Flow Matching (CFM) extends this to learned trajectories conditioned on observations.
% GFM leverages these advances by applying CFM to neural weight trajectories, treating optimization as a conditional dynamical system whose vector field is shaped by the optimizer and observed weights. This approach bridges the gap between flow-based generative modeling and neural network training dynamics.

% \paragraph{Time Series Forecasting and Surrogates.}
% Recent results in time-series prediction have shown that simple linear models often rival deep sequence models for long-horizon forecasting~\citep{zeng2023dlinear}. Inspired by this, LFD-2 used minimal state representations for forecasting, showing surprisingly strong performance. In contrast, Transformer and RNN models are often impractical for direct application to weight sequences due to high dimensionality.
\paragraph{Time Series Forecasting.}
Our work is related to long-term time series forecasting, where both classical and modern models have been extensively explored. Recent studies have demonstrated that simple linear models can rival deep sequence models for long-horizon prediction, as shown in DLinear and its variants~\citep{zeng2023transformers}. Yet applying standard time-series models such as ~\citep{ nie2022time} or RNNs directly to neural weight sequences is often impractical due to the high dimensionality and nonstationary nature of the optimization trajectory. While recent efforts explore foundation models for time series~\citep{ye2024survey}, their computational demands remain a challenge in our setting. %Moreover, transformations designed to capture periodicity and seasonal trends~\citep{li2023revisiting} are unlikely to benefit weight modeling, as training trajectories lack such temporal regularities. In our context, feedforward models offer a more scalable alternative, though they may underutilize the algorithmic structure that governs weight evolution.

% \paragraph{Learning Curve Extrapolation.}
% Another related  or orthogonal line of research focuses on predicting the evolution of training loss or accuracy over time instead of weight trajectory , often to facilitate early stopping or performance estimation. Prior work on learning curve extrapolation has employed diverse approaches, including Bayesian models~\citep{klein2017learning} and historical learning trends~\citep{gargiani2019predicting}. And the use of architecture-based performance prediction has leveraged graph neural networks (GNNs) to estimate the final accuracy of unseen architectures, for example \citep{ding2025architecture}. 

\paragraph{Learning Curve Extrapolation.}
An orthogonal line of research focuses on predicting the evolution of training loss or accuracy over time, rather than modeling the underlying weight trajectories. These approaches are often used to enable early stopping or to estimate final model performance. Prior work on learning curve extrapolation includes Bayesian methods~\citep{klein2017learning} as well as techniques based on historical learning trends~\citep{gargiani2019probabilistic}. In a related direction, architecture-based performance prediction has leveraged graph neural networks to estimate the final accuracy of unseen architectures~\citep{ding2025architecture}.

\section{Preliminary}
\label{prelim} 
% \paragraph{Gradient Descent.} First-order gradient descent methods have been the primary workhorse for optimizing machine and deep learning models. The typical goal is to minimize a designated loss function $L$ with respect to certain parameters $\mathbf{w}$ within a defined parameter space. The iterative update for the parameters using gradient descent is: $\mathbf{w}_{i+1} \leftarrow \mathbf{w}_i - \alpha_i \nabla L(\mathbf{w}_i; D)$ for a given dataset $D$ with an initial $\mathbf{w}_0$ by choice of initialization methods. For a large scale deep learning systems in language and vision, these parameters are more commonly called weights (and bias). We use paramters and weights interchangeably in our setting. A vanilla mini-batch stochastic gradient descent (SGD) performs iterative update by a sampled subset from $D$, i.e. $\mathbf{w}_{i+1} \leftarrow \mathbf{w}_i - \alpha_i \nabla L(\mathbf{w}_i; B)$ where $B \sim p(D)$.  
\paragraph{Gradient Descent.}
First-order gradient descent methods are the foundational tools for optimizing machine learning and deep learning models. The objective is to minimize a loss function $L$ with respect to parameters $\mathbf{w}$ over a given parameter space. The standard update rule is given by: $\mathbf{w}_{i+1} \leftarrow \mathbf{w}_i - \alpha_i \nabla L(\mathbf{w}_i; D)$, where $D$ denotes the dataset and $\alpha_i$ is the learning rate at step $i$. The process begins from an initial point $\mathbf{w}_0$, typically determined by a chosen initialization scheme. In large-scale deep learning systems—particularly in language and vision tasks—these parameters are commonly referred to as weights (and biases); we use the terms parameters and weights interchangeably in our context. In practice, a more scalable variant, mini-batch stochastic gradient descent (SGD), updates the parameters using a randomly sampled subset $B$ of the dataset, i.e., $\mathbf{w}_{i+1} \leftarrow \mathbf{w}_i - \alpha_i \nabla L(\mathbf{w}_i; B)$ where $B \sim p(D)$.

% \paragraph{Flow Matching.} 
% Flow Matching (FM) is a framework for learning a probability density path $p_t(x)$ by matching a learnable vector field $\mathbf{v}_t(x)$ that governs the dynamics of a flow $\phi_t(x)$ modeled by ODE $\frac{d}{dt} \phi_t(x) = \mathbf{v}_t(\phi_t(x))$, to a target vector field $\mathbf{u}_t(x)$, which generates the probability flow. %Given an initial distribution $p_0$ (e.g., a standard normal distribution) and a target distribution $p_1$ that approximates an unknown data distribution $q(x)$, the goal of FM is to construct a probability path $p_t$ that smoothly transitions from $p_0$ to $p_1$. 
% The (unconditional) Flow Matching can be done by regressing a neural network $\mathbf{v}_\theta (x,t) $ minimizing the MSE loss :
% \begin{equation}
% \mathcal{L}_{FM}(\theta) = \mathbb{E}_{t, p_t(x)} \| \mathbf{v}_\theta(x,t) - \mathbf{u}_t(x) \|^2,
% \end{equation}
% A typical example is in generative images where $\mathbf{u}_t(x)$ is the target vector field generating the image density path $p_t(x)$ from source to target. The expectation is taken over time $t \sim \mathcal{U}[0,1]$ and samples $x \sim p_t(x)$. In practice, the unconditional flow matching loss and its conditional part $\mathcal{L}_{FM}(\theta)$ differ by a constant which leads to effective minimizing the more tractable conditional loss. \cite{flowsanddiffusions2025}
\paragraph{Flow Matching.}
Flow Matching is a framework for learning a probability path $\{p_t(x)\}_{t \in [0,1]}$ by training a learnable vector field $\mathbf{v}_t(x)$ to match a target vector field $\mathbf{u}_t(x)$ that governs the evolution of a probability flow. The underlying dynamics are modeled by an ordinary differential equation (ODE), $\frac{d}{dt} \phi_t(x) = \mathbf{v}_t(\phi_t(x))$, where $\phi_t(x)$ denotes the trajectory of a sample $x$ over time. The goal is to learn $\mathbf{v}_\theta(x, t)$ such that it closely approximates $\mathbf{u}_t(x)$ across the path defined by the distribution $p_t(x)$. In the unconditional setting, this is achieved by minimizing the mean squared error between the learnable and target vector fields:
\begin{equation}
\mathcal{L}_{\text{FM}}(\theta) = \mathbb{E}_{t,\, x \sim p_t(x)} \left[ \| \mathbf{v}_\theta(x,t) - \mathbf{u}_t(x) \|^2 \right],
\end{equation}
where $t \sim \mathcal{U}[0,1]$ is sampled uniformly over time. A typical use case is in generative modeling, where $\mathbf{u}_t(x)$ transports samples from a base distribution (e.g., Gaussian noise) to a target data distribution along the path $p_t(x)$. In practice, the conditional version of FM often provides a more tractable objective, differing from the unconditional loss by a constant~\citep{flowsanddiffusions2025}.

% \section{Method}
% \paragraph{Problem statement}. Deep learning applications are generally interested in the following setting:
% given a sequence of training weights $\{\mathbf{w}_i\}_{i=0}^n$ where $n \ge 0$, we are interested in forecasting some steps to a convergent weight $\mathbf{w}_*$.  For ease of notation, we will take $m$ steps such as $\mathbf{w}_m$ with very small $\epsilon$ distance from $\mathbf{w}_*$. 

% we develop a flow matching framework for (S)GD which aims to find the vector field that governs the dynamics of probability path $p_t(\mathbf{w})$ conditioned on 
% $\{\mathbf{w}_i\}_{i=0}^n$. This can be formulated by the follow equation.
% \begin{equation}
% \label{eqn:obj}
% \mathcal{L}_{CFM}(\theta) = \mathbb{E}_{t, p_t(\mathbf{w})} \| \mathbf{v}_\theta(\mathbf{w},t) - \mathbf{u}_t(\mathbf{w}| \mathbf{w}_0, \mathbf{w}_1,...,\mathbf{w}_n, \mathbf{w}_m) \|^2,
% \end{equation}
\section{Method}
\begin{figure}[htbp]
    \centering
    \includegraphics[width=\linewidth]{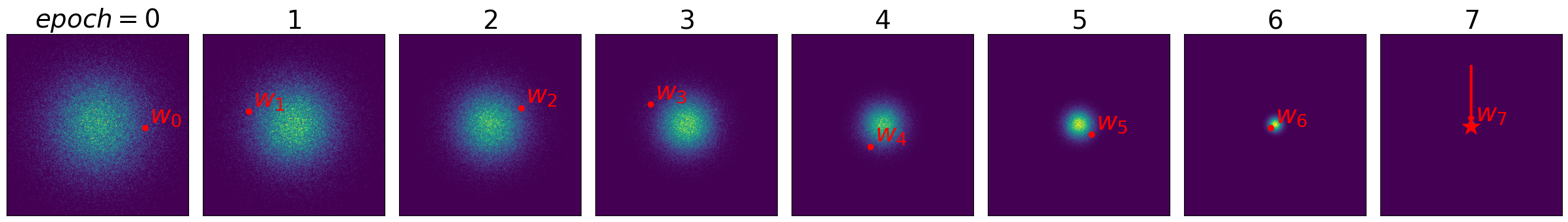}
    \caption{
    Visualization of the evolving weight distribution over training epochs. Each panel shows the probability density $p_t(\mathbf{w})$ at a given epoch $t$, with red dots indicating the actual weight positions $\mathbf{w}_0, \dots, \mathbf{w}_7$. As training progresses, weights move from a broad initialization distribution toward a more concentrated region near convergence. The final state $\mathbf{w}_7$ approximates the optimal weight $\mathbf{w}_*$.}
    \label{fig:weight_path}
\end{figure}

\paragraph{Problem Statement.}
In typical deep learning applications, we are given a partial sequence of training weights $\{\mathbf{w}_i\}_{i=0}^n$ generated by an optimization algorithm such as stochastic gradient descent. Our objective is to forecast a future weight $\mathbf{w}_*$ that approximates the converged or optimal solution. However, the true dynamics of gradient-based optimization are generally not analytically tractable or directly observable. As a practical alternative, we aim to forecast a later-stage weight $\mathbf{w}_m$ that lies within a small neighborhood of the converged point. To this end, we develop a flow matching framework tailored for optimization dynamics. Our approach learns a vector field that governs the evolution of a probability path $p_t(\mathbf{w})$ over weight space, conditioned on the observed prefix $\{\mathbf{w}_i\}_{i=0}^n$. This objective is formalized as:
\begin{equation}
\label{eqn:obj}
\mathcal{L}_{\text{CFM}}(\theta) = \mathbb{E}_{t,\, \mathbf{w} \sim p_t(\mathbf{w})} \left\| \mathbf{v}_\theta(\mathbf{w}, t) - \mathbf{u}_t\left(\mathbf{w} \mid \mathbf{w}_0, \dots, \mathbf{w}_n, \mathbf{w}_m \right) \right\|^2,
\end{equation}
where $\mathbf{v}_\theta$ is a neural network parameterizing the learned vector field, and $\mathbf{u}_t$ is the (implicit) target vector field that drives the trajectory toward convergence. An example of the evolving density path $p_t(\mathbf{w})$ is illustrated in Figure~\ref{fig:weight_path}.

\subsection{Reparameterization for Continuous Flow}
We focus on the problem of modeling weight evolution in neural networks trained with first-order optimizers such as SGD, Adam, and RMSProp—common choices in deep learning applications. For notational simplicity, we omit the dataset $D$ (or mini-batch $B$) from our formulation and reintroduce it as needed. The weight update rule for SGD can be written as:
\begin{equation}
    \mathbf{w}_{i+1} - \mathbf{w}_i = -\alpha_i \nabla L(\mathbf{w}_i),
\end{equation}
or equivalently, $\Delta \mathbf{w}_i = -\alpha_i \nabla L(\mathbf{w}_i)$, which resembles a discretized ordinary differential equation (ODE) with a step size of 1. However, flow matching is inherently defined in continuous time, modeling a smooth probability density path $p_t(\mathbf{w})$ for $t \in [0,1]$. To bridge the gap between discrete updates and continuous flows, we reinterpret the weight sequence $\{\mathbf{w}_i\}_{i=0}^m$ as samples from a continuous trajectory $\mathbf{w}(t)$. Under this perspective, the SGD update rule corresponds to an Euler discretization of a continuous-time ODE. Taking the limit as the step size approaches zero yields:
\begin{equation}
\frac{d}{dt} \mathbf{w}(t) = -\alpha(t) \nabla L(\mathbf{w}(t)), \quad t \in [0,1],
\end{equation}
with boundary conditions $\mathbf{w}(0) = \mathbf{w}_0$ and $\mathbf{w}(1) = \mathbf{w}_m$, and where $\alpha(t)$ denotes a continuous interpolation of the discrete learning rates $\alpha_i$.

Since the original trajectory is indexed by discrete steps $i = 0, 1, \dots, m$, we introduce a normalized continuous-time parameterization: $t = \frac{i}{m}$, such that $\mathbf{w}(t)$ denotes a continuous approximation of the weight sequence. At intermediate values of $t$, we define $\mathbf{w}(t)$ via linear interpolation between two adjacent weights:
\begin{equation}
\mathbf{w}(t) = \text{Interp}(\mathbf{w}_{\lfloor tm \rfloor}, \mathbf{w}_{\lfloor tm \rfloor + 1}) = (1 - \omega)\, \mathbf{w}_{\lfloor tm \rfloor} + \omega\, \mathbf{w}_{\lfloor tm \rfloor + 1},   
\end{equation}
where $\omega = tm - \lfloor tm \rfloor$. This construction enables a smooth transition from discrete updates to a continuous-time flow, compatible with the flow matching objective. Throughout the remainder of this paper, we use parenthetical notation $\mathbf{w}(t)$ for continuous trajectories and subscript notation $\mathbf{w}_i$ for their discrete counterparts.

\subsection{Conditional Flow Matching with Initial \& Optimal Weight}
We begin with the simplified setting where only the initial and final weights are available, i.e., $n = 0$. In this case, the conditional target vector field $\mathbf{u}_t(\mathbf{w} \mid \mathbf{w}_0, \dots, \mathbf{w}_n, \mathbf{w}_m)$ reduces to $\mathbf{u}_t(\mathbf{w} \mid \mathbf{w}_0, \mathbf{w}_m)$. For simplicity, we assume the weights are initialized from an isotropic Gaussian distribution, aligning with the noise assumptions commonly used in flow matching frameworks. While we adopt Gaussian initialization for theoretical convenience, modern deep learning models often employ more structured initialization schemes. For example, Xavier (Glorot) initialization~\citep{glorot2010understanding} normalizes weights based on layer fan-in and fan-out to maintain gradient stability, while Kaiming initialization~\citep{he2015delving} extends this to ReLU-based architectures. These strategies could, in principle, be incorporated by adjusting the prior distribution $p_0(\mathbf{w})$ to match their scaling characteristics.

In this two-point setting, we can apply a conditional optimal transport formulation analogous to those used in generative modeling~\citep{lipmanflow, tong2023conditional}. Without access to intermediate gradients or weight values, the most direct estimate of the target vector field is the displacement between the final and initial weights: $\mathbf{u}_t = \mathbf{w}_m - \mathbf{w}_0$. This leads to the following conditional flow matching loss from Equation \ref{eqn:obj} :
\begin{equation}
\mathcal{L}_{\text{CFM}}(\theta) = \mathbb{E}_{t \sim \mathcal{U}[0,1],\, \mathbf{w} \sim p_t(\mathbf{w})} \left\| \mathbf{v}_\theta(\mathbf{w}, t) - (\mathbf{w}_m - \mathbf{w}_0) \right\|^2.
\end{equation}
The corresponding conditional path $p_t(\mathbf{w} \mid \mathbf{w}_0, \mathbf{w}_m)$ is modeled as a linear Gaussian interpolation: $p_t(\mathbf{w} \mid \mathbf{w}_0, \mathbf{w}_m) = \mathcal{N}(t \mathbf{w}_m + (1 - t) \mathbf{w}_0, \sigma^2 \mathbf{I}),
$ with $\sigma^2 \to 0$ in the deterministic limit.

It is worth noting that this setup does not require observing any actual weight updates—it simply learns to interpolate between $\mathbf{w}_0$ and $\mathbf{w}_m$. While this may be limited in practical utility for real-world forecasting, it offers theoretical insight. Specifically, from the telescoping identity: $\mathbf{w}_m - \mathbf{w}_0 = \sum_{i=0}^{m-1} \alpha_i \nabla L(\mathbf{w}_i)$, we can interpret the model as learning a vector field that approximates the learning-rate-weighted average of the gradient field over the trajectory. This establishes a foundation for more expressive conditional models using richer prefixes.

\subsection{Conditional Flow Matching with Intermediate Weights}

We now consider the more general setting where we have access to several training steps, i.e., $n > 0$. With a continuous-time relaxation, if both the learning rate schedule $\alpha(t)$ and gradient $\nabla L(\mathbf{w}(t))$ were available for all $t$, we could directly apply flow matching by training a learnable vector field $\mathbf{v}_\theta(\mathbf{w}, t)$ to approximate the true dynamics:
\begin{equation}
\mathcal{L}_{\text{CFM}}(\theta) = \mathbb{E}_{t \sim \mathcal{U}[0,1],\, \mathbf{w} \sim p_t(\mathbf{w})} \left\| \mathbf{v}_\theta(\mathbf{w}, t) + \alpha(t) \nabla L(\mathbf{w}(t)) \right\|^2.
\end{equation}
While the learning rate $\alpha(t)$ is typically known from a schedule, gradients are not stored during training and are thus unavailable at inference or forecasting time. To address this, we propose a surrogate target vector field using finite differences over the interpolated trajectory:
\begin{equation}
\mathcal{L}_{\text{CFM}}(\theta) = \mathbb{E}_{t \sim \mathcal{U}[0,1],\, \mathbf{w} \sim p_t(\mathbf{w})} \left\| \mathbf{v}_\theta(\mathbf{w}, t) - \left(\mathbf{w}(t + \tfrac{1}{m}) - \mathbf{w}(t)\right) \right\|^2.
\end{equation}

We consider two cases:

\textbf{(i) Short-horizon forecasting ($m = n+1$):} In this case, we are predicting just one step ahead. The finite difference $(\mathbf{w}(t + \tfrac{1}{m}) - \mathbf{w}(t))$ approximates the underlying gradient dynamics, and the flow field effectively models the next-step direction. Since we already observe weights up to $\mathbf{w}_{n+1} = \mathbf{w}_m$, this setup is somewhat trivial: forecasting becomes unnecessary when $\mathbf{w}_{m-1}$ is already close to $\mathbf{w}_m$.

\textbf{(ii) Multi-step forecasting ($m > n+1$):} In this more realistic and useful setting, we forecast several steps ahead. To do so, we use observed gradients (approximated by finite differences) for $t = \tfrac{i}{m}$ where $i \in \{0, \dots, n\}$, and apply linear interpolation beyond that, i.e., for $i \in \{n+1, \dots, m\}$. This hybrid strategy captures two objectives: (a) preserving faithfulness to the observed prefix $\{\mathbf{w}_i\}_{i=0}^n$, and (b) guiding the model toward an optimal endpoint via conditional extrapolation from $\mathbf{w}_n$ to $\mathbf{w}_m$.

We balance these objectives with a soft transition using indicator-weighted interpolation. Let $Z := \mathbbm{1}(t < \tfrac{n}{m})$ denote whether a time point falls within the observed prefix. We then define our overall flow matching training objective as:
\begin{align}
\mathcal{L}_{\text{CFM}}(\theta) = \mathbb{E}_{t \sim \mathcal{U}[0,1],\, \mathbf{w} \sim p_t(\mathbf{w})} \big( 
& \beta Z \left\| \mathbf{v}_\theta(\mathbf{w}, t) - \left(\mathbf{w}(t + \tfrac{1}{m}) - \mathbf{w}(t)\right) \right\|^2 \nonumber \\
& + \gamma (1 - Z) \left\| \mathbf{v}_\theta(\mathbf{w}, t) - \left(\mathbf{w}_m - \mathbf{w}_n\right) \right\|^2 \big).
\label{eqn:cfm}
\end{align}

Here, $\beta$ and $\gamma$ control the trade-off between prefix faithfulness and optimal extrapolation. These weights can also be made time-dependent—e.g., by applying exponential decay to emphasize recent steps in the observed prefix. Alternatively, a cutoff strategy can be used, focusing the regularization on the last $k$ steps prior to $\mathbf{w}_n$. This formulation enables training of a flow-matching model that can generalize beyond observed weights to forecast long-horizon optimization trajectories. %A prototype algorithm is provided to illustrate how our objective can be applied within stochastic gradient descent. In practice, the same logic can be extended to mini-batches of variable-length sequences by sampling interpolation times $t$ independently for each instance.
\paragraph{Midpoint Integration with Forecast Consistency Penalty.}
Beyond conditional flow matching loss in Equation \ref{eqn:cfm}, we introduce a consistency regularization term based on midpoint integration. Specifically, we use a second-order midpoint method to evolve the flow field from the observed weight $\mathbf{w}_n$ to a predicted terminal weight $\hat{\mathbf{w}}_m$. This prediction is computed as: $\hat{\mathbf{w}}_m = \mathbf{w}_n + \Delta t \cdot \mathbf{v}_\theta\left(\mathbf{w}_n + \tfrac{1}{2} \Delta t \cdot \mathbf{v}_\theta(\mathbf{w}_n, t_n), t_n + \tfrac{1}{2} \Delta t\right)$
where $\Delta t = t_m - t_n$, and the vector field $\mathbf{v}_\theta$ is evaluated at the midpoint in both time and space. To ensure that the predicted final weight $\hat{\mathbf{w}}_m$ remains accurate, we introduce a least squares penalty between the prediction and the true weight $ \mathcal{L}_{\text{Pred}} = \mathbf{w}_m$ (i.e. $\|\hat{\mathbf{w}}_m - \mathbf{w}_m\|^2$). The total loss becomes $ \mathcal{L}_{\text{GFM}} = \mathcal{L}_{\text{CFM}} + \zeta \cdot \mathcal{L}_{\text{Pred}}$ where $\zeta$ is a hyperparameter controlling the strength of this consistency regularization. This term encourages the learned flow to not only match local vector fields but also produce globally accurate forecasts under integration. A prototype training algorithm incorporating this objective is outlined in Algorithm~\ref{alg:tr}. 

\begin{algorithm}[t]
\caption{Mini-batch Flow Matching for Optimal Weight Forecasting}
\label{alg:tr}
\begin{algorithmic}[1]
\Require Dataset $\{s_i = \{\mathbf{w}_0, \dots, \mathbf{w}_n, \mathbf{w}_m\}\}_{i=1}^N$, batch size $B$, vector field $v_\theta(\mathbf{w}, t)$, $\beta$, $\gamma$, $\zeta$
\Ensure Trained network $v_\theta$
\For{each mini-batch $\{s_j\}_{j=1}^B$}
    \State Sample $t \sim \mathcal{U}[0,1]$, $\omega \gets tm - \lfloor tm \rfloor$, $Z \gets \mathbbm{1}(t < \tfrac{n}{m})$
    \State $\mathbf{w}(t) \gets Z \cdot \text{Interp}(\mathbf{w}_{\lfloor tm \rfloor}, \mathbf{w}_{\lfloor tm \rfloor + 1}, \omega) + (1-Z)(t \mathbf{w}_m + (1-t) \mathbf{w}_0)$
    \State $\mathbf{v}_{\text{target}} \gets Z \cdot (\mathbf{w}_{\lfloor tm \rfloor + 1} - \mathbf{w}_{\lfloor tm \rfloor}) + (1-Z)(\mathbf{w}_m - \mathbf{w}_n)$
    \State $\mathcal{L}_{\text{CFM}} \gets \| v_\theta(\mathbf{w}(t), t) - \mathbf{v}_{\text{target}} \|^2 \cdot (\beta Z + \gamma (1 - Z))$
    \State $\hat{\mathbf{w}}_{\text{mid}} \gets \mathbf{w}_n + \tfrac{1}{2}(1 - \tfrac{n}{m}) \cdot v_\theta(\mathbf{w}_n, \tfrac{n}{m})$
    \State $\hat{\mathbf{w}}_m \gets \mathbf{w}_n + (1 - \tfrac{n}{m}) \cdot v_\theta(\hat{\mathbf{w}}_{\text{mid}}, \tfrac{n}{m} + \tfrac{1}{2}(1 - \tfrac{n}{m}))$
    \State $\mathcal{L}_{\text{GFM}} \gets \mathcal{L}_{\text{CFM}} + \zeta \cdot \| \hat{\mathbf{w}}_m - \mathbf{w}_m \|^2$
    \State Update $\theta \gets \theta - \eta \nabla_\theta \mathcal{L}_{\text{GFM}}$
\EndFor
\end{algorithmic}
\end{algorithm}

\subsection{Inference Procedure}
At inference time, we are given a discrete sequence of observed parameters $\{\mathbf{w}_0, \dots, \mathbf{w}_n\}$, obtained from training or prior observations. The objective is to forecast a future weight $\mathbf{w}_m$ beyond the observed sequence by leveraging the learned flow model $v_\theta(\mathbf{w}, t)$. Starting from the last known state $\mathbf{w}_n$, we simulate the continuous-time dynamics prescribed by $v_\theta$ through numerical integration (e.g., Euler's method). Specifically, the parameters are iteratively updated as $\mathbf{w}_{k+1} = \mathbf{w}_k + h \cdot v_\theta(\mathbf{w}_k, t_k)$,

where $h$ is the integration step size, and $t_k$ progresses toward the terminal time $t=1$. To ensure computational efficiency, we introduce an early stopping criterion based on the magnitude of updates. The integration halts when $\|\Delta\mathbf{w}_k\| < \tau$, where $\Delta\mathbf{w}_k = h \cdot v_\theta(\mathbf{w}_k, t_k)$, and $\tau$ is a user-defined tolerance threshold, typically set to $10^{-6}$. If convergence is detected before reaching $t = 1$, the final output $\mathbf{w}_m$ is set to the corresponding converged parameter. Otherwise, if the maximum number of integration steps is reached without satisfying the stopping condition, $\mathbf{w}_m$ is set to the state at $t = 1$.

\subsection{Optimizer Dynamics}
\paragraph{Momentum-based Optimizers.}
Our framework naturally extends to optimizers with momentum-like dynamics. For example, stochastic gradient descent with momentum~\citep{sutskever2013importance} updates parameters as $\mathbf{w}_{i+1} = \mathbf{w}_i - \alpha_i \mathbf{m}_i$, where $\mathbf{m}_i = \mu \mathbf{m}_{i-1} + (1 - \mu) \nabla L(\mathbf{w}_i)$ for some decay factor $\mu$. Unrolling this recursion yields $\mathbf{w}_{i+1} - \mathbf{w}_i = -\alpha_i (1 - \mu) \sum_{k=0}^{i} \mu^k \nabla L(\mathbf{w}_{i - k})$, which shows how updates depend on a weighted history of past gradients. Our learned vector field $v_\theta(\mathbf{w}, t)$ approximates these dynamics by leveraging the universal function approximation capability of neural networks~\citep{hornik1989multilayer}.
\paragraph{Adaptive Optimizers.}
Similarly, our approach generalizes to adaptive methods such as RMSProp~\citep{heaton2018ian}, where parameters are updated as $\mathbf{w}_{i+1} = \mathbf{w}_i - \alpha_i \cdot \frac{\mathbf{g}_i}{\sqrt{\mathbf{v}_i} + \epsilon}$, with gradients $\mathbf{g}_i = \nabla L(\mathbf{w}_i)$ and accumulated squared gradients $\mathbf{v}_i = \mu \mathbf{v}_{i-1} + (1 - \mu) (\mathbf{g}_i \circ \mathbf{g}_i)$. Expanding $\mathbf{v}_i$ reveals its dependency on past squared gradients, resulting in the update $\mathbf{w}_{i+1} - \mathbf{w}_i = -\alpha_i \cdot \frac{\mathbf{g}_i}{\sqrt{(1 - \mu) \sum_{k=0}^{i} \mu^k (\mathbf{g}_{i - k} \circ \mathbf{g}_{i - k})} + \epsilon}$. The $\epsilon$ term ensures numerical stability by preventing division by small values. Although our vector field $v_\theta$ does not explicitly model these second-moment accumulations, its expressiveness allows it to implicitly capture such adaptive behaviors through training on weight trajectories. Moreover, hybrid optimizers like Adam~\citep{kingma2014adam}, which combine momentum and adaptive scaling, can also be effectively modeled within this framework.

\section{Experiments}
% \paragraph{Baseline Models.} 
% We compare our proposed model, \textbf{GFM}, against six representative baselines: DLinear, LFD-2, Introspection, WNN, Transformer, and LSTM. These baselines span a range of modeling paradigms, from classical weight forecasting methods to modern deep learning architectures. 
% \textbf{Introspection}~\cite{sinha2017introspection} and \textbf{WNN}~\cite{jang2023learning} are specifically designed for weight forecasting, learning to predict future weights directly from historical parameter sequences. 
% \textbf{DLinear}~\cite{zeng2023transformers} is a recent time series forecasting model that questions the necessity of complex Transformer-based designs; it instead relies on linear projections and RevIN normalization~\cite{kim2021reversible} to model temporal and channel-wise patterns. 
% We also include \textbf{Transformer} and \textbf{LSTM} baselines of comparable size (on the order of $10^3$ parameters), adapted for our task of forecasting optimization trajectories. 
% Finally, we adapt \textbf{LFD-2}~\cite{shou2025less}, a recent efficient single-layer feedforward model that predicts the final converged weight using only the initial and early-stage parameters, $w_0$ and $w_n$.
\paragraph{Baseline Models.}
We compare \textbf{GFM} against six representative baselines: DLinear, LFD-2, Introspection, WNN, Transformer, and LSTM. \textbf{Introspection}~\cite{sinha2017introspection} and \textbf{WNN}~\cite{jang2023learning} are designed for short-term weight forecasting. %, learning to predict future parameters from historical sequences. 
\textbf{DLinear}~\cite{zeng2023transformers} applies linear projections with RevIN normalization~\cite{kim2021reversible} %, emphasizing simplicity for time series forecasting. 
We also adapt \textbf{Transformer} and \textbf{LSTM} models with $\sim 10^3$ parameters for weight trajectory prediction. Lastly, \textbf{LFD-2}~\cite{shou2025less} predicts final weights using only initial and early-stage parameters ($\mathbf{w}_0$, $\mathbf{w}_n$), providing an efficient feedforward baseline. More details around models are included in Appendix \ref{sec:appendix-baseline}.

\subsection{Synthetic Optimization Trajectories}

\paragraph{Task Data Generation.} To evaluate the ability of models to forecast converged weights under varying target functions, we construct a synthetic dataset based on gradient-based optimization of linear regression models. Each trajectory captures the evolution of model parameters $\mathbf{w}_i = (w_i, b_i) \in \mathbb{R}^2$ over time during training. Each regression task is defined by a unique ground-truth linear function, with coefficients sampled as:
$a \sim \mathcal{N}(2.0, 0.01), \quad b \sim \mathcal{N}(1.0, 0.01)$. Given each sampled pair $(a, b)$, we generate a dataset of 100 input-output pairs with inputs $x \sim \text{Uniform}([-1, 1])$ and additive noise $\varepsilon \sim \mathcal{N}(0, \sigma^2)$. The process along with the following trajectory generation is detailed in Appendix \ref{sec:appendix-synthetic-lr}.

\paragraph{Trajectory Generation.} For each task, we train a two-parameter linear model using one of five optimizers: SGD, Adam, RMSprop, AdamW, or Adagrad. The model is initialized identically across tasks and optimized using mean squared error loss. Parameters are recorded over $T = 199$ update steps, resulting in a trajectory of length 200 that converges to task-specific optima. For each optimizer, we generate $N = 50$ such trajectories and repeat the process across 5 random seeds.

\paragraph{Forecasting Final Weights.} We evaluate the ability of forecasting models to predict the final converged weight given only a prefix of the optimization trajectory. Each dataset is split into 30 training and 20 testing sequences. All models are trained with mean squared error between the predicted and true final weights used as the evaluation metric. Each model is conditioned on the first $n{+}1$ updates, $\{\mathbf{w}_0, \mathbf{w}_1, \ldots, \mathbf{w}_n\}$ where $n = 4$, and tasked with forecasting the final converged weight $\mathbf{w}_m$ at step $m = 199$. This controlled setting provides a benchmark for evaluating how well flow-based models capture the underlying optimization dynamics and generalize to diverse convergence targets.

\begin{figure}[htb]
    \centering
    \includegraphics[width=0.86\textwidth]{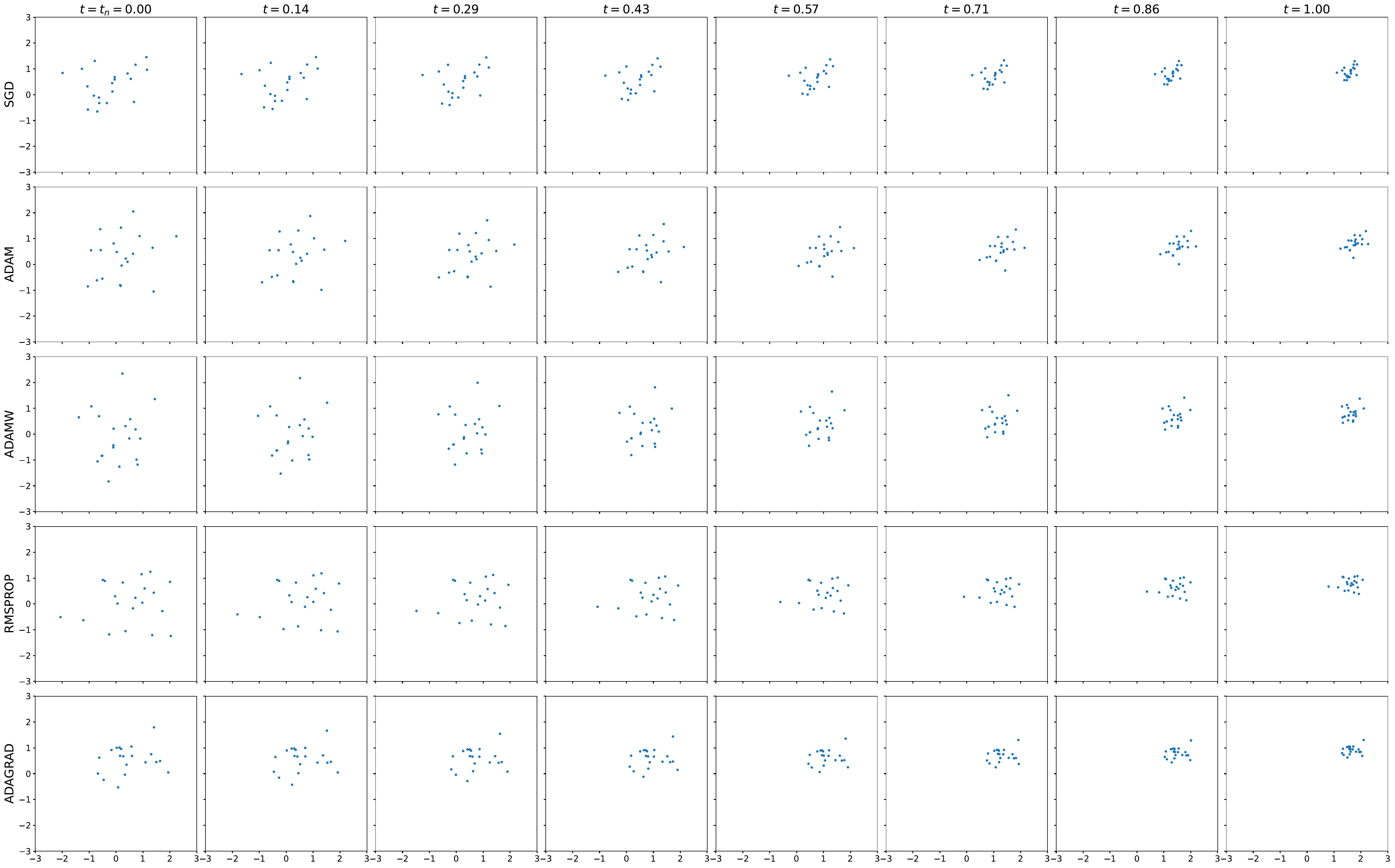}
    \caption{
        Forecasted optimization trajectories produced by GFM for different optimizers (rows) and time steps (columns), shown for 20 test trajectories starting from $t_n$. Each row visualizes the evolution of weights from initialization to convergence, conditioned on the first five steps. GFM successfully learns smooth transitions and captures the distinct dynamics characteristic of each optimizer.
    }
    \label{fig:gfm_inference_grid}
\end{figure}

\paragraph{Results.} Table~\ref{tab:final-column-wise-best} reports test MSE across five optimizers and seven models, including our proposed model, GFM with default hyperparameters $\beta =1.0$, $\gamma=1.0$, $\zeta=100.0$. A discussion on these paramters are in Appendix \ref{sec:param-sen}. Across all optimizers, GFM consistently achieves either the lowest or second-lowest error, indicating strong generalization performance. In particular, GFM attains the best MSE under \texttt{adagrad}, \texttt{adamw}, and \texttt{rmsprop}, and remains highly competitive under \texttt{adam} and \texttt{sgd}, closely matching the Transformer baseline. Compared to the strongest alternative (Transformer), GFM offers comparable accuracy with simpler architecture. All other baselines—including DLinear, and WNN—exhibit significantly higher errors, especially under adaptive optimizers. These results underscore the effectiveness of GFM in learning the underlying optimization dynamics and forecasting converged weights accurately. Our proposed model, GFM, adopts a generative perspective that captures the first-order update behavior inherent in gradient-based optimization. This is demonstrated in Figure~\ref{fig:gfm_inference_grid}, which visualizes inferred optimization trajectories across five different optimizers. The smooth and directed transitions observed in the plots reflect GFM's ability to emulate the underlying update dynamics and generate convergent parameter flows. As expected, the final parameter distributions tend to cluster around the known ground-truth optimum, which follows a Gaussian distribution $\mathcal{N}([2.0, 1.0], 0.01 \cdot \mathbf{I})$. These visualizations highlight GFM’s capacity to generalize across diverse optimizer-induced trajectories while maintaining stability and convergence toward the target solution.

% \begin{table}[htbp]
% \centering
% \caption{Test MSE with std (formatted as mean(std)) across models and optimizers. Bold indicates the best, italic the second best.}
% \resizebox{0.8\textwidth}{!}{%
% \begin{tabular}{lccccc}
% \toprule
% Model/Optimizer & SGD & Adam & AdamW & RMSProp & Adagrad \\
% \midrule
% Introspection  &  .2299 (.1333) &  .2759 (.1416) &  .2706 (.1393) &  .1175 (.0783) &  .1342 (.1037) \\
% DLinear     &  .0228 (.0076) &  .5171 (.1863) &  .4896 (.1764) &  .3848 (.1391) &  .3145 (.0905) \\
% LFD2        &  2.427 (.526) &  2.682 (.582) & 2.653 (.576) &  2.201 (.458) &  2.125 (.405) \\
% WNN         &  .8114 (.3536) &  .9348 (.2746) & .9102 (.2696) &  .5393 (.2266) &  .4603 (.1545) \\
% LSTM        &  1.601 (.029) &  1.591 (.023) & 1.572 (.024) &  1.567 (.032) &  1.525 (.049) \\
% Transformer &  \textbf{.0203 (.0067)} &  \textbf{.0196 (.0056)} & \textit{.0192 (.0054)} &  \textit{.0204 (.0072)} &  \textit{.0155 (.0042)} \\
% GFM (ours)  &  \textit{.0207 (.0131)} &  \textit{.0201 (.0119)} & \textbf{.0162 (.0067)} &  \textbf{.0190 (.0111)} &  \textbf{.0134 (.0023)} \\
% \bottomrule
% \end{tabular}
% }
% \label{tab:final-column-wise-best}
% \end{table}

\begin{table}[htbp]
\centering
\caption{Test MSE with std (formatted as mean(std)) across models and optimizers. Bold indicates the best, italic the second best.}
\resizebox{0.8\textwidth}{!}{%
\begin{tabular}{lccccc}
\toprule
Model/Optimizer & SGD & Adam & AdamW & RMSProp & Adagrad \\
\midrule
Introspection  &  .230 (.133) &  .276 (.142) &  .271 (.139) &  .118 (.078) &  .134 (.104) \\
DLinear        &  .023 (.008) &  .517 (.186) &  .490 (.176) &  .385 (.139) &  .315 (.091) \\
LFD2           &  2.427 (.526) &  2.682 (.582) & 2.653 (.576) &  2.201 (.458) &  2.125 (.405) \\
WNN            &  .811 (.354) &  .935 (.275) &  .910 (.270) &  .539 (.227) &  .460 (.155) \\
LSTM           &  1.601 (.029) &  1.591 (.023) & 1.572 (.024) &  1.567 (.032) &  1.525 (.049) \\
Transformer    &  \textbf{.020 (.007)} &  \textbf{.020 (.006)} & \textit{.019 (.005)} &  \textit{.020 (.007)} &  \textit{.016 (.004)} \\
GFM (ours)     &  \textit{.021 (.013)} &  \textbf{.020 (.012)} & \textbf{.016 (.007)} &  \textbf{.019 (.011)} &  \textbf{.013 (.002)} \\
\bottomrule
\end{tabular}
}
\label{tab:final-column-wise-best}
\end{table}

\subsection{Generalization Across MLP Architectures}

We evaluate the generalization ability of our proposed model, GFM, by training on one neural architecture and forecasting the optimization trajectory of another. Specifically, we generate training trajectories on the same task data as in linear regression using four widely adopted optimizers in deep learning: SGD, Adam, AdamW, and RMSprop, each with a learning rate of 0.001 over 199 epochs. All models are initialized with weights drawn from a standard normal distribution. The training architecture is a 3-layer MLP with hidden sizes [2, 2, 1], and the test architecture is a 2-layer MLP with hidden sizes [4, 1]. For each training run, we record the full optimization trajectory (including initialization), resulting in tensors of shape $(50, 200, 15)$. During inference, GFM observes the early portion of a test trajectory up to a forecast point $n$ (e.g., $n = 4$) and predicts the final converged weights at epoch 199. More details are included in Appendix \ref{sec:dg-mlp}. 

\begin{figure}[thbp]
    \centering
    \begin{subfigure}[t]{0.245\textwidth}
        \centering
        \includegraphics[width=\textwidth]{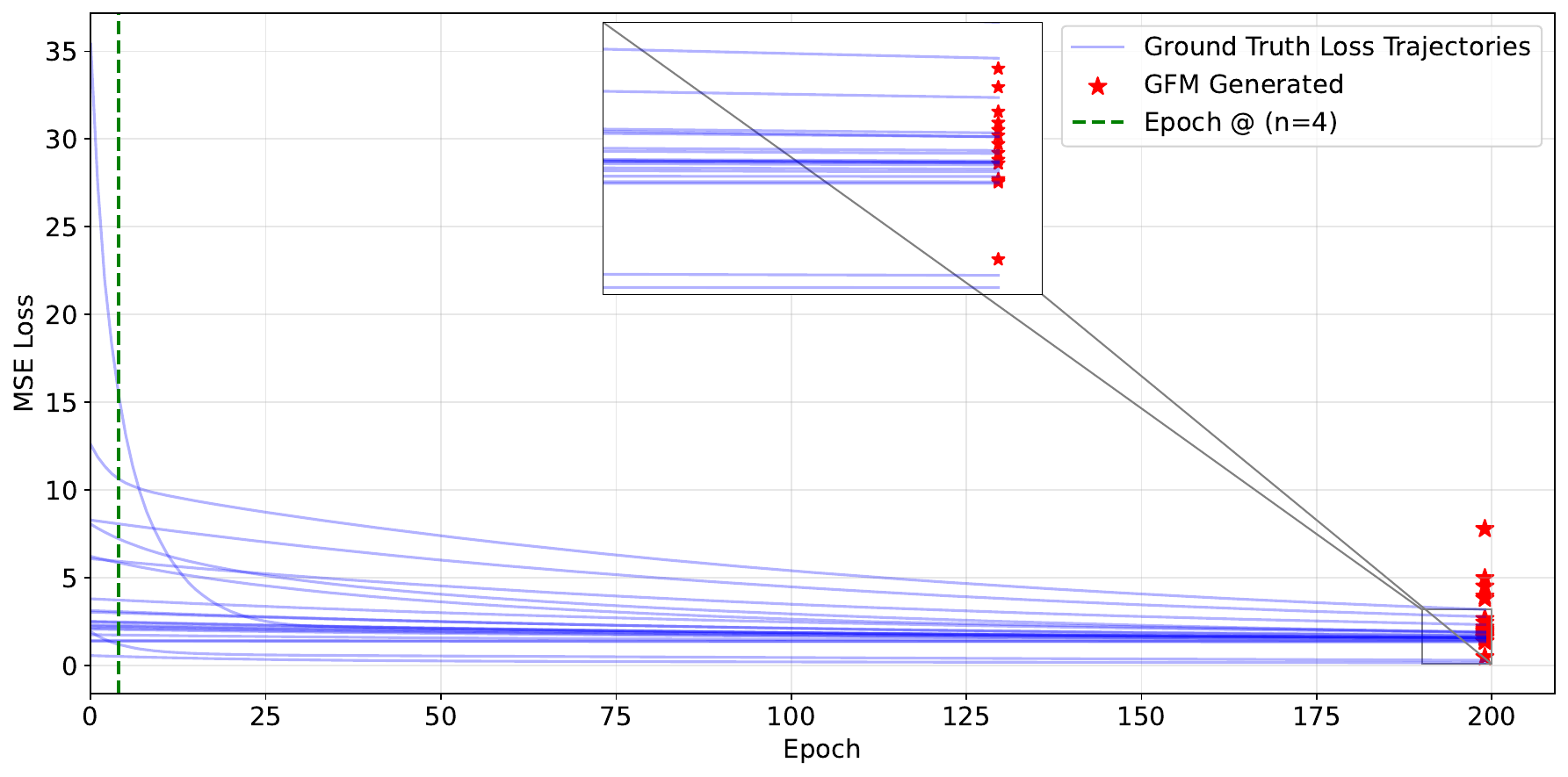}
        \caption{SGD}
    \end{subfigure}
    \hfill
    \begin{subfigure}[t]{0.245\textwidth}
        \centering
        \includegraphics[width=\textwidth]{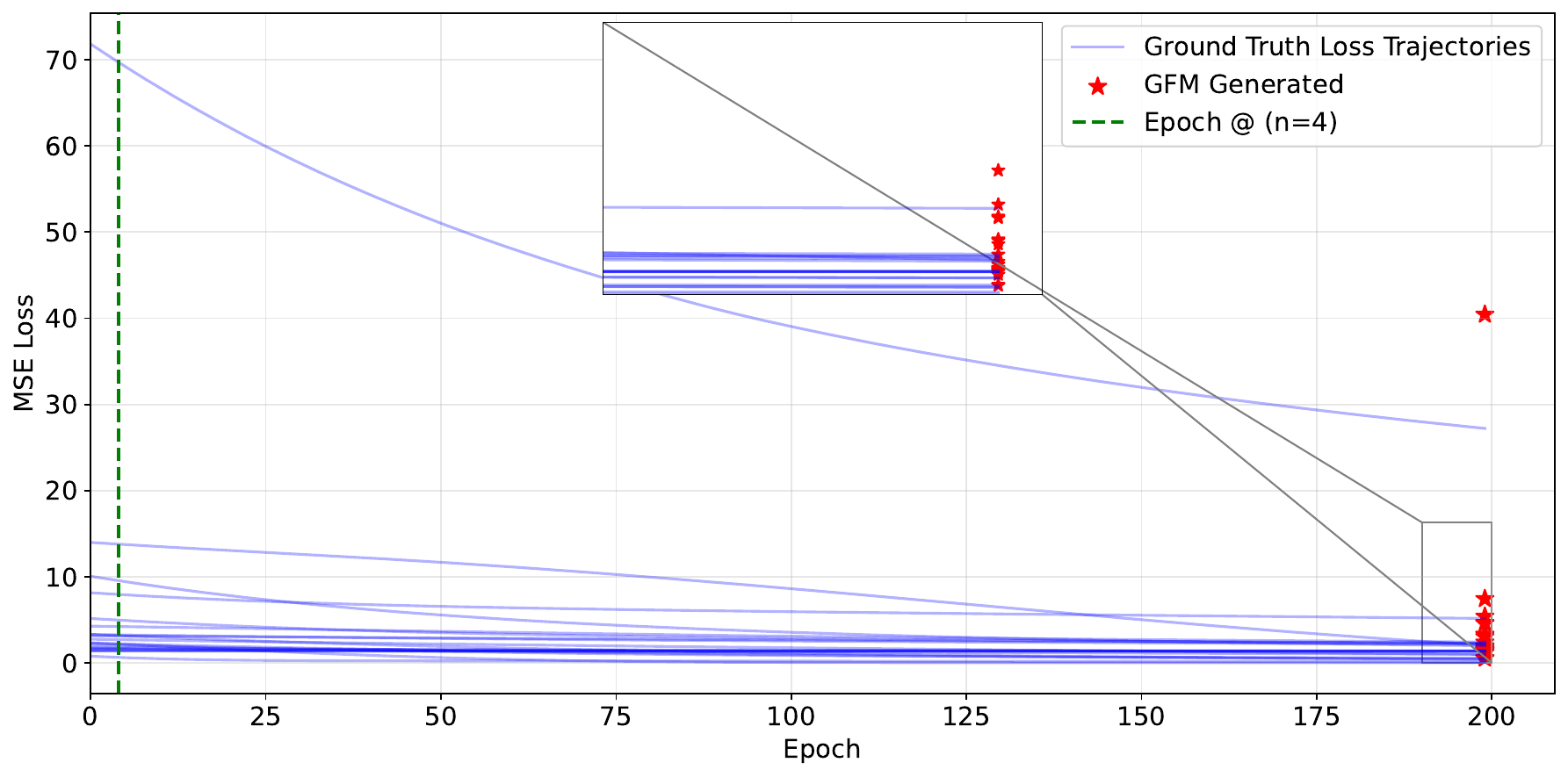}
        \caption{Adam}
    \end{subfigure}
    \hfill
    \begin{subfigure}[t]{0.245\textwidth}
        \centering
        \includegraphics[width=\textwidth]{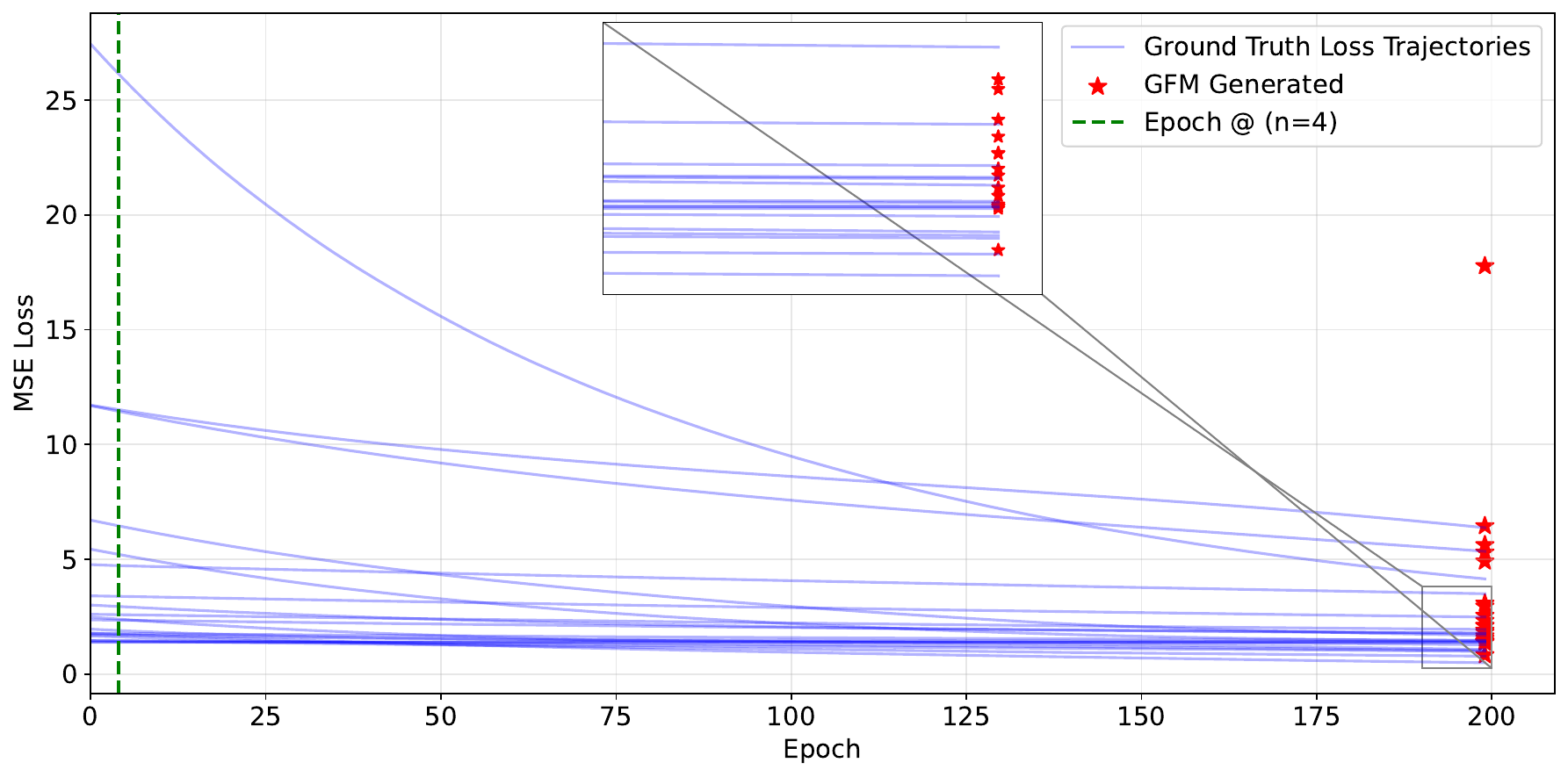}
        \caption{AdamW}
    \end{subfigure}
    \hfill
    \begin{subfigure}[t]{0.245\textwidth}
        \centering
        \includegraphics[width=\textwidth]{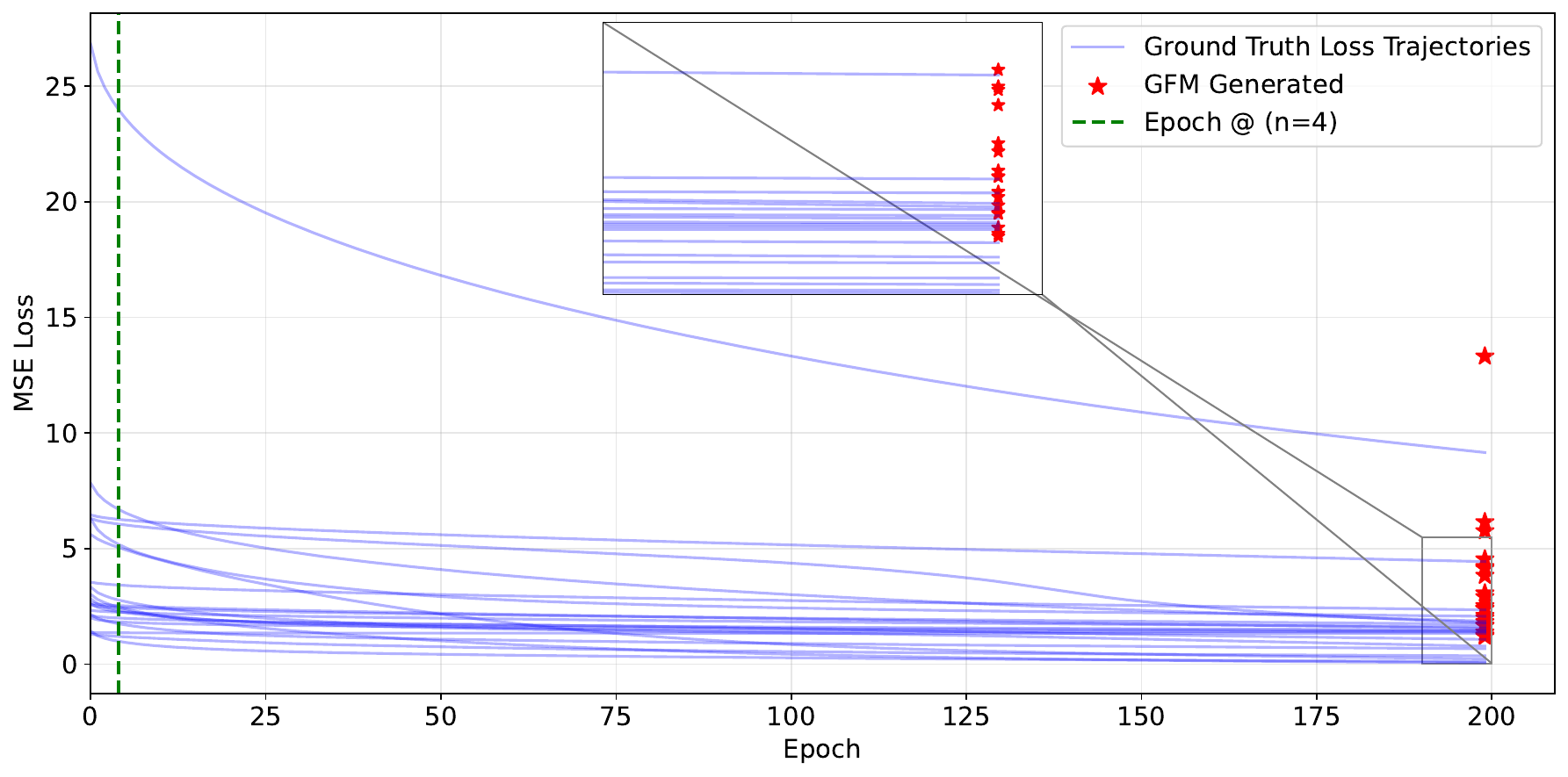}
        \caption{RMSprop}
    \end{subfigure}
    
    \caption{Loss trajectories of 2-layer MLPs trained with different optimizers. Each blue curve shows task-wise training loss over epochs, and red stars mark flow-matched predictions.}
    \label{fig:loss-trajectories-4opt}
\end{figure}

We then evaluate the predicted weights by measuring the resulting test loss and comparing it to the ground-truth task loss trajectory. Figure~\ref{fig:loss-trajectories-4opt} shows the predicted vs. ground-truth loss for each optimizer at forecast step $n=4$. Notably, GFM-generated weights often yield comparable, and occasionally better, loss than the true converged weights. We observe similar trends for earlier forecast steps ($n=1,2,3$). While we match parameter counts for ease of comparison, this framework naturally extends to settings with mismatched parameter sizes via padding mechanism. 

\subsection{Application: Training Dynamics Forecasting on CIFAR-10}
\paragraph{Trajectory Generation.} 
% We construct three fixed discrete architecture spaces for CNNs and Transformers.  
We evaluate GFM on CNN and Transformer architectures trained on CIFAR-10. 
For CNNs, we implement the NAS-Bench201 search space by enumerating cells with 4 nodes, where each edge selects one of 5 candidate operations (none, skip-connect, 1×1 conv, 3×3 conv, 3×3 avg-pool), following \cite{dong2020bench}.  
For Transformers, we sample layer counts in $ \{1,2,3,4\}$, hidden sizes in $\{64,128,256,512,1024\}$, and dropout $\in\{0.0, 0.1, 0.3\}$.  
Each layer is a self-attention plus MLP block (including LayerNorm and residual connections).  
For each architecture type, we generate 100 training trajectories using SGD over 200 epochs on CIFAR-10, varying random initialization seeds to ensure diversity and split the 100 trajectories into 60 train and 40 test sequences. Note that the ground truth weights may not converge or reach optimality within the 200-epoch training period, due to the inherent challenges of this real-world dataset. Details on data preprocessing and training setup are provided in Appendix~\ref{sec:appendix-nn-generation}.

\paragraph{Forecasting Setup and Evaluation.} 
We evaluate models that predict final weights $\mathbf{w}_{199}$ based on early training snapshots ${\mathbf{w}_0,\ldots,\mathbf{w}_n}$, with $n=4$. We report (1) test MSE between predicted and true weights and (2) the cross-entropy loss of the predicted weights on the source classification task.%, denoted as $f_{\text{source}}$. %Additionally, we report 
%$f^{\text{gt}}_{\text{source}}$, 
%the source-task loss at the true final weights.%For the CNN dataset, batch size is 5; for Transformer, it is 1. All forecasting models are trained using the AdamW optimizer. We repeat each experiment with five different random seeds and report the mean and standard deviation across runs. 

% \paragraph{Results.} 
% As shown in Table~\ref{tab:application}, on the Transformer dataset, GFM achieves the best performance with an MSE of 0.043 and a source-task loss $f_{\text{source}}$ of 1.117, representing a 49.8\% reduction in $f_{\text{source}}$ compared to the second-best method (LFD2, $f_{\text{source}} = 2.226$). This demonstrates the model’s strength in accurately forecasting final weights that generalize well on the original task.
% In contrast, for the CNN dataset, GFM does not achieve the lowest MSE or $f_{\text{source}}$. While WNN yields the smallest MSE (0.365), it  results in a poorer $f_{\text{source}}$ (3.605) compared to introspection ($f_{\text{source}} = 2.368$), suggesting that MSE alone does not fully capture the quality of predicted weights in terms of task performance. A possible reasoning is that models may overfit to the weight trajectory’s Euclidean structure without capturing functionally meaningful features relevant to downstream accuracy. In some cases, lower MSE might not necessarily imply better functional generalization, underscoring the need to evaluate forecasting methods using both geometric and functional metrics.
\paragraph{Results.}
% Table~\ref{tab:application} shows that GFM achieves the best $f_{\text{source}}$ on Transformers (1.117), reducing error by 49.8\% compared to LFD-2 (2.226), demonstrating superior forecasting of functionally meaningful weights. On CNNs, GFM's MSE is competitive but lower $f_{\text{source}}$ is achieved by Introspection (2.369). Notably, WNN attains the smallest MSE (0.365) but with poor task loss (3.605), indicating that MSE alone may fail to capture downstream generalization for real-world highly nonconvex problems. These results highlight the importance of evaluating both geometric (MSE) and functional (task loss) metrics when assessing weight forecasting quality.
Table~\ref{tab:application} shows that GFM achieves the best alignment with downstream task performance on Transformers, with an $f_{\text{source}}$ of 1.117, representing a 49.8\% improvement over LFD-2 the second best performing model. This suggests that GFM produces weight forecasts that are more effective for the target task. On CNNs, while GFM's MSE remains competitive, Introspection achieves a better task alignment. Interestingly, WNN attains the smallest MSE (0.365), but its higher $f_{\text{source}}$ (3.605) corresponds to poorer task accuracy, indicating that lower prediction error in parameter space does not always translate into better task performance. These findings emphasize the importance of evaluating forecasting methods by both their prediction error and their impact on final task outcomes.

\begin{table}[htbp]
\centering
\caption{Forecasting performance on CIFAR-10 (mean (std) over 5 seeds). %Ground-truth task loss: 1.014 (CNN), 0.041 (Transformer). 
DLinear/WNN omitted for Transformers due to memory issues.}
\resizebox{0.73\textwidth}{!}{%
\begin{tabular}{lcc|cc}
\toprule
\multicolumn{1}{c}{} & \multicolumn{2}{c}{\textbf{CNN}} & \multicolumn{2}{c}{\textbf{Transformer}} \\
\cmidrule(lr){2-3} \cmidrule(lr){4-5}
Model & MSE & $f_{\text{source}}$  & MSE & $f_{\text{source}}$ \\
\midrule
Introspection & 0.421 (0.000) & \textbf{2.369} (0.003) & 0.054 (0.000) & 2.225 (0.005) \\
DLinear       & 0.900 (0.123) & 309.361 (126.423) & -- & -- \\
LFD2          & 0.421 (0.000) & 2.381 (0.007) & 0.054 (0.000) & 2.226 (0.002) \\
WNN           & \textbf{0.365} (0.000) & 3.605 (0.245) & -- & -- \\
LSTM          & 0.466 (0.004) & 2.604 (0.024) & 0.054 (0.000) & 2.266 (0.052) \\
Transformer   & 0.543 (0.013) & 3.526 (0.387) & 0.053 (0.000) & 2.228 (0.004) \\
GFM           & 0.485 (0.021) & 3.369 (0.416) & \textbf{0.043} (0.000) & \textbf{1.117} (0.001) \\
\bottomrule
\end{tabular}
}
\label{tab:application}
\end{table}

\subsection{Ablation Experiments}

% \paragraph{Effect of Different Initializers.}
% We evaluate the impact of weight initialization on forecasting performance  for synthetic trajectories from linear regression .Since our ne is linear without a ReLU activation function, Xavier initialization is theoretically appropriate and Kaiming is not. To test this empirically, we compare Xavier Uniform and Xavier Normal across various optimizers on our synthetic datasets with 5 repeats. From Table \ref{tab:xavier_init_results}, across most optimizers, Xavier Normal leads to consistently lower forecasting error compared to Xavier Uniform and better than normally initialized results in Table \ref{tab:final-column-wise-best}. This suggests that using Xavier Normal may be a better default choice in linear settings, especially when forecasting optimization trajectories.

\paragraph{Effect of Different Initializers.}
\begin{wraptable}{r}{0.6\textwidth}
    \centering
    \vspace{-0.6em}
    \caption{MSE loss for different optimizers under Xavier initialization with $n=4$. Arrows indicate performance relative to default Normal.}
    \resizebox{0.6\textwidth}{!}{%
    \begin{tabular}{lccccc}
    \toprule
    Optimizer & SGD & Adam & AdamW & RMSprop & Adagrad \\
    \midrule
    Xavier Normal  & .020 (.014) $\downarrow$ & .017 (.008) $\downarrow$ & .017 (.008) $\uparrow$ & .016 (.009) $\downarrow$ & .013 (.003) $\downarrow$ \\
    Xavier Uniform & .021 (.008) $\uparrow$  & .026 (.007) $\uparrow$  & .026 (.007) $\uparrow$  & .019 (.006) $\uparrow$  & .017 (.005) $\uparrow$ \\
    \bottomrule
    \end{tabular}
    }
    \label{tab:xavier_init_results}
\end{wraptable}

We assess how weight initialization affects forecasting performance on synthetic linear regression trajectories. Since our network is linear without ReLU activations, Xavier initialization is theoretically suitable, while Kaiming is not. Empirically, we compare Xavier Uniform and Xavier Normal across optimizers (5 repeats). As shown in Table~\ref{tab:xavier_init_results}, Xavier Normal consistently achieves lower forecasting errors than Xavier Uniform and improves over default normal initialization (Table~\ref{tab:final-column-wise-best}). This suggests Xavier Normal as a better default for linear settings when forecasting optimization trajectories.

\paragraph{Forecasting from Initialization Only.}
\begin{wraptable}{r}{0.55\textwidth}
    \centering
    \vspace{-0.1em}
    \caption{MSE loss for $n = 0$. $\uparrow$ indicates a slight increase compared to $n = 4$.}
    \resizebox{0.55\textwidth}{!}{%
    \begin{tabular}{lccccc}
    \toprule
    Optimizer & SGD & Adam & AdamW & RMSprop & Adagrad \\
    \midrule
    GFM & .023 (.008) $\uparrow$ & .025 (.012) $\uparrow$ & .025 (.012) $\uparrow$ & .025 (.010) $\uparrow$ & .025 (.014) $\uparrow$ \\
    \bottomrule
    \end{tabular}
    }
    \label{tab:gfm_optimizer_compact}
\end{wraptable}

We examine whether GFM can forecast final weights using only initialization ($n = 0$) on synthetic linear regression trajectories. As shown in Table~\ref{tab:gfm_optimizer_compact}, GFM achieves reasonable accuracy despite lacking intermediate observations, demonstrating its ability to infer optimization dynamics solely from initial states.

\section{Conclusion}
We presented \emph{Gradient Flow Matching} (GFM), a continuous-time framework that models neural network training as optimizer-aware vector fields. By explicitly incorporating gradient-based dynamics, GFM outperforms sequence modeling baselines while maintaining simplicity and generality across architectures and optimizers. A potential \textbf{limitation} is GFM relies on observed weight sequences and does not explicitly model second order curvature information, which may limit its extrapolation accuracy in highly non-stationary training regimes. Future work includes extending GFM to integrate richer contextual signals, such as gradient statistics or architecture-conditioned priors. %to further improve its forecasting capability and generalization.
% We introduced \emph{Gradient Flow Matching} (GFM), a continuous-time framework that forecasts final weights by modeling optimizer-aware dynamics. GFM outperforms sequence baselines while remaining simple and general across architectures. Its reliance on observed weight sequences, without explicit curvature modeling, may limit extrapolation in non-stationary regimes. Future work includes enriching GFM with gradient statistics and architecture priors to enhance generalization.

% \begin{ack}
% Use unnumbered first level headings for the acknowledgments. All acknowledgments

% Do {\bf not} include this section in the anonymized submission, only in the final paper. You can use the \texttt{ack} environment provided in the style file to automatically hide this section in the anonymized submission.
% \end{ack}

\section*{References}
% \bibliography{ref}
% % \bibliographystyle{unsrt}
% \bibliographystyle{unsrtnat}  % ✅ numeric & unsorted, works well with natbib

% References follow the acknowledgments in the camera-ready paper. Use unnumbered first-level heading for
% the references. Any choice of citation style is acceptable as long as you are
% consistent. It is permissible to reduce the font size to \verb+small+ (9 point)
% when listing the references.
% Note that the Reference section does not count towards the page limit.
% \medskip

% {
% \small

% [1] Alexander, J.A.\ \& Mozer, M.C.\ (1995) Template-based algorithms for
% connectionist rule extraction. In G.\ Tesauro, D.S.\ Touretzky and T.K.\ Leen
% (eds.), {\it Advances in Neural Information Processing Systems 7},
% pp.\ 609--616. Cambridge, MA: MIT Press.

% [2] Bower, J.M.\ \& Beeman, D.\ (1995) {\it The Book of GENESIS: Exploring
%   Realistic Neural Models with the GEneral NEural SImulation System.}  New York:
% TELOS/Springer--Verlag.

% [3] Hasselmo, M.E., Schnell, E.\ \& Barkai, E.\ (1995) Dynamics of learning and
% recall at excitatory recurrent synapses and cholinergic modulation in rat
% hippocampal region CA3. {\it Journal of Neuroscience} {\bf 15}(7):5249-5262.
% }

\newpage
%%%%%%%%%%%%%%%%%%%%%%%%%%%%%%%%%%%%%%%%%%%%%%%%%%%%%%%%%%%%

\appendix

\section{Baseline Models}
\label{sec:appendix-baseline}
\paragraph{Baseline Models.}
We compare our proposed method against several representative baselines for weight trajectory forecasting:

\textbf{Introspection}~\cite{sinha2017introspection} and \textbf{WNN}~\cite{jang2023learning} are specialized models designed to predict future network weights from historical sequences. Both directly regress future weights but differ in architectural complexity; WNN employs weight-difference features and regularization schemes, while Introspection uses a shallow feedforward predictor.

\textbf{DLinear}~\cite{zeng2023transformers} is a recent time series model emphasizing simplicity, relying on separate temporal and channel-wise linear projections with RevIN normalization~\cite{kim2021reversible}. Despite its origin in forecasting tasks, we adapt it for weight trajectory prediction.

We further include \textbf{Transformer} and \textbf{LSTM} baselines, implemented with comparable model capacities. These models process sequences of past weights and predict the final weight configuration via sequence-to-one regression. 

Lastly, \textbf{LFD-2}~\cite{shou2025less} represents an efficient forecasting approach that extrapolates final weights from only the initial and early-stage parameters ($\mathbf{w}_0$ and $\mathbf{w}_n$) using a lightweight feedforward network. A summary of model configurations are shown in Table \ref{tab:baseline-comparison}.

For fair comparison, all models are trained using the Adam optimizer with a learning rate of $0.0001$ and evaluated by mean squared error (MSE) between predicted and true final weights $\mathbf{w}_{199}$. Models observe prefix sequences up to $t = 4$ and are trained using a 60/40 train/test split. Batch sizes of 16 or 32 are selected based on best performance. All experiments are conducted on a MacBook Pro equipped with an Apple M3 Max chip and 48GB memory. This unified setup ensures consistent evaluation protocols and a level playing field across all forecasting methods.

\begin{table}[htbp]
\centering
\caption{Baseline architectures and configurations. Parameter counts are approximate and depend on input dimension $D$. All models forecast $\mathbf{w}_{199}$ from prefix steps up to $t=4$.}
\resizebox{\textwidth}{!}{%
\begin{tabular}{lcccc}
\toprule
\textbf{Model} & \textbf{\# Layers} & \textbf{\# Params} & \textbf{Input Shape} & \textbf{Key Operations} \\
\midrule
Introspection  & 2 (Linear, ReLU, Linear) & $\approx 4D \cdot 100 + 100 \cdot D$ & $[N, D, 4]$ & Fully-connected regressor \\
DLinear        & 2 Temporal + 2 Channel MLP + RevIN & $\approx 4D \cdot d_{model}$ & $[N, t+1, D]$ & Linear projections, normalization \\
LFD-2          & 1 Linear layer & $2D \cdot D$ & $[N, D, 2]$ & Extrapolation from $w_0$, $w_n$ \\
WNN            & $\sim$12 Dense + Calculus layers & $\approx 20K$ & $[X1, dX]$ & Weight-difference features, L1 regularization \\
Transformer    & Embed + 1 EncoderLayer + Decoder & $\approx D \cdot 32 + 32 \cdot D$ & $[N, t+1, D]$ & Self-attention, feedforward layers \\
LSTM           & 2 LSTM layers + FC & $\approx 4D \cdot 16 + 16 \cdot D$ & $[N, t+1, D]$ & Sequence modeling with LSTM memory \\
GFM (Ours)     & 4-layer MLP Vector Field: (Linear + 3xELU + Linear) & $(D+1) \cdot 64 + 64^2 \cdot 2 + 64 \cdot D$ & $[N, D]$ for weights + $t$ & Conditional flow field modeling \citep{lipman2024flow} \\
\bottomrule
\end{tabular}
}
\label{tab:baseline-comparison}
\end{table}

% \begin{table}[htbp]
% \centering
% \caption{Training configurations for baseline models. All methods forecast $\mathbf{w}_{199}$ from prefix steps up to $t=4$, evaluated via MSE on a 60/40 train/test split.}
% \resizebox{0.6\textwidth}{!}{%
% \begin{tabular}{lcccccc}
% \toprule
% \textbf{Model} & \textbf{Opt.} & \textbf{LR} & \textbf{Batch} & \textbf{Epochs} & \textbf{Prefix ($t$)} & \textbf{Split} \\
% \midrule
% Introspection & Adam & 0.0001 & 16, 32 & 10000 & 4 & 60/40 \\
% DLinear       & Adam & 0.0001 & 16, 32 & 10000 & 4 & 60/40 \\
% LFD-2         & Adam & 0.0001 & 16, 32 & 10000 & 4 & 60/40 \\
% WNN           & Adam & 0.0001 & 16, 32 & 1000 & 4 & 60/40 \\
% Transformer   & Adam & 0.0001 & 16, 32 & 1000 & 4 & 60/40 \\
% LSTM          & Adam & 0.0001 & 16, 32 & 1000 & 4 & 60/40 \\
% GFM (Ours)    & Adam & 0.0001 & 16, 32 & 1000 & 4 & 60/40 \\
% \bottomrule
% \end{tabular}
% }
% \label{tab:training-configs}
% \end{table}

\section{Synthetic Linear Regression Data Generation}
\label{sec:appendix-synthetic-lr}
To evaluate our forecasting models in a controlled setting, we generate synthetic optimization trajectories for linear regression tasks using gradient-based optimizers. Specifically, we simulate the weight evolution of a single-layer linear model trained via stochastic optimization, tracking the parameter updates over time.

\paragraph{Task Setup.} 
We consider scalar linear regression tasks where each dataset is defined by a ground-truth slope and intercept. For each generated task, we sample input points $x \in [-1, 1]$ and add Gaussian noise. The ground-truth slope and intercept are drawn from Gaussian distributions (mean 2.0, std 0.1), depending on the dataset configuration.

\paragraph{Trajectory Generation.} 
For each task, we optimize the model using one of several optimizers (SGD, Adam, RMSProp, AdamW, Adagrad) over $199$ training steps. Weight initialization follows either Xavier uniform, Xavier normal, or standard Gaussian distributions. During training, we record the full weight trajectory by saving the concatenated slope and intercept parameters at each step. 

\paragraph{Optimizer Configuration.}
We adopt optimizer-specific hyperparameters for trajectory generation: a learning rate of $0.01$ for SGD, Adam, RMSProp, and AdamW, and $0.1$ for Adagrad. RMSProp uses a smoothing constant $\alpha=0.99$ and weight decay $0.01$ by default. For Adam-based methods, we employ standard beta values $(0.9, 0.999)$.

\paragraph{Dataset Construction.}
For each optimizer, we generate five trajectory datasets using random seeds $\{0, 1, 2, 3, 4\}$ to ensure diversity. Each dataset contains $100$ distinct optimization trajectories. The resulting data is saved in a structured directory format for reproducibility.

\begin{figure}[htbp]
    \centering
    \begin{subfigure}[t]{0.19\textwidth}
        \includegraphics[width=\linewidth]{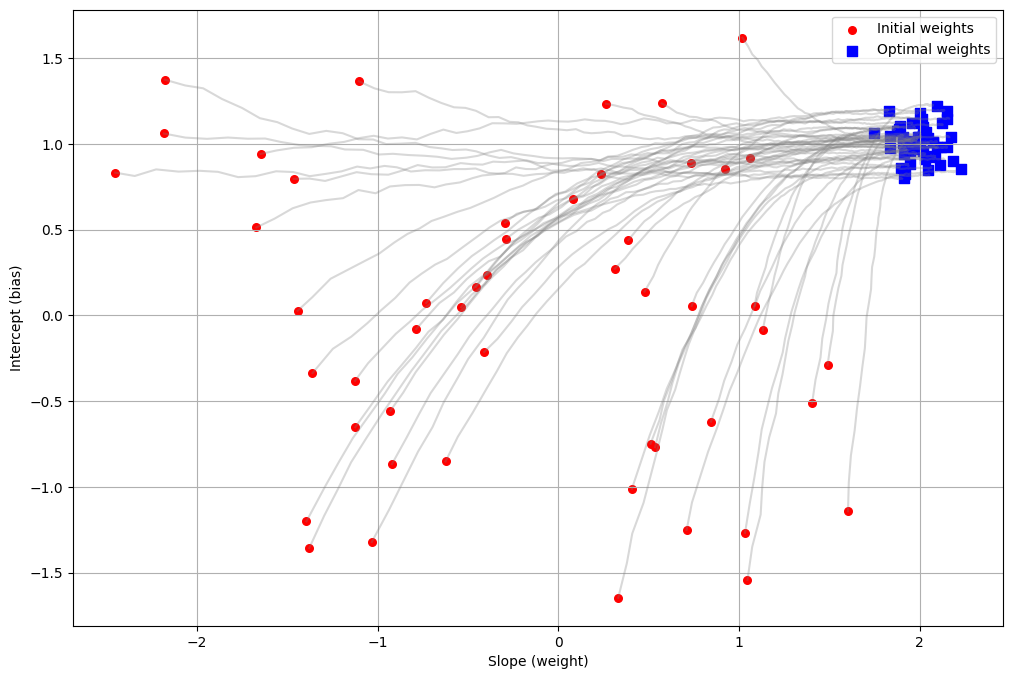}
        \caption{SGD}
    \end{subfigure}
    \hfill
    \begin{subfigure}[t]{0.19\textwidth}
        \includegraphics[width=\linewidth]{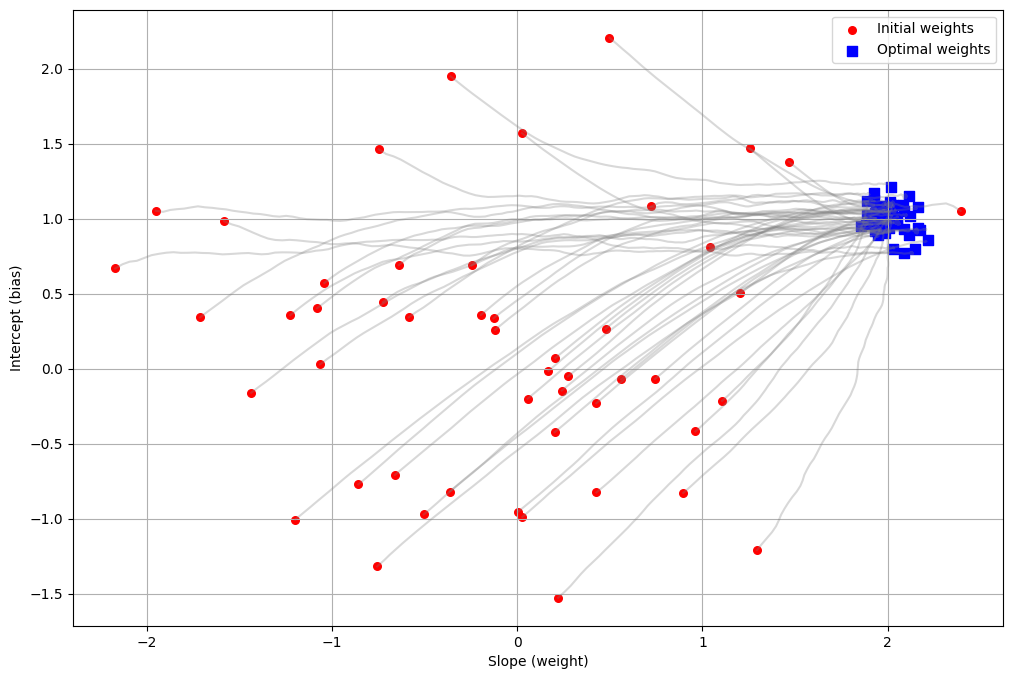}
        \caption{Adam}
    \end{subfigure}
    \hfill
    \begin{subfigure}[t]{0.19\textwidth}
        \includegraphics[width=\linewidth]{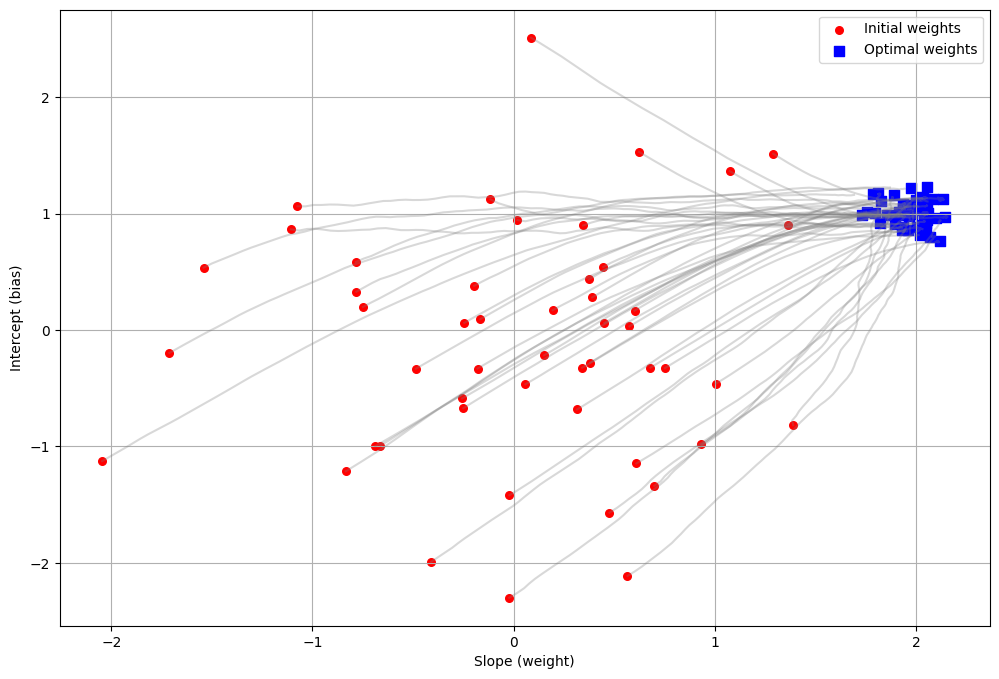}
        \caption{AdamW}
    \end{subfigure}
    \hfill
    \begin{subfigure}[t]{0.19\textwidth}
        \includegraphics[width=\linewidth]{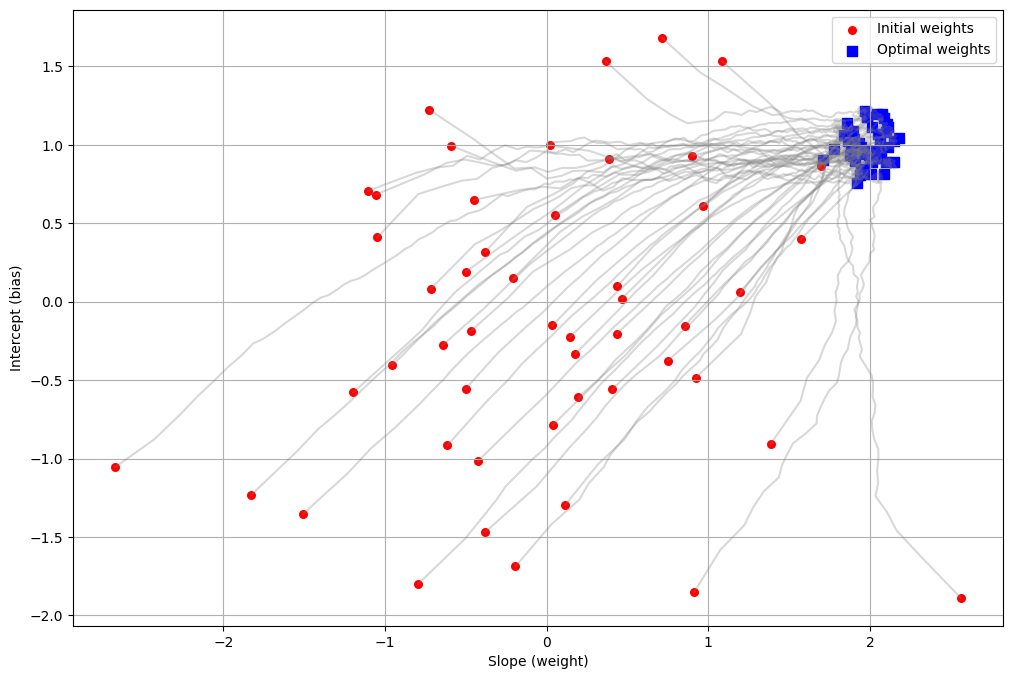}
        \caption{RMSprop}
    \end{subfigure}
    \hfill
    \begin{subfigure}[t]{0.19\textwidth}
        \includegraphics[width=\linewidth]{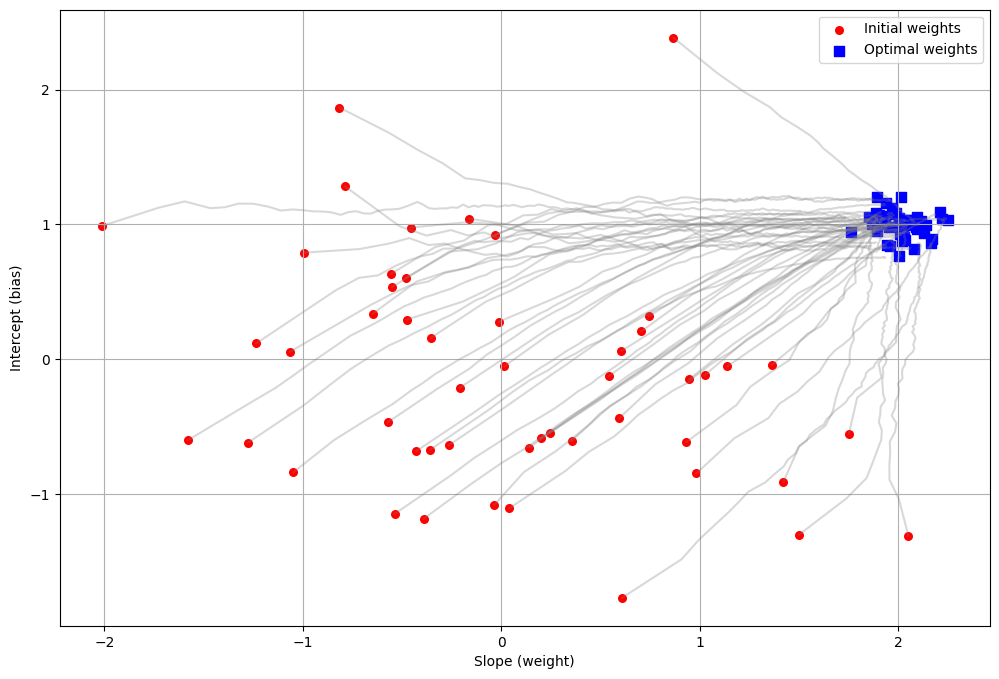}
        \caption{Adagrad}
    \end{subfigure}
    \caption{Optimization trajectories of weight parameters from initialization to convergence for five optimizers. Each plot visualizes 50 trajectories for linear regression tasks under Gaussian weight sampling.}
    \label{fig:optimizer-trajectories}
\end{figure}

\paragraph{Visualization.}
Figure~\ref{fig:optimizer-trajectories} illustrates sample weight trajectories, showcasing the evolution of slope and intercept parameters toward convergence under 5 optimizers with varying seed initializations. 

\paragraph{Output Format.}
Each generated dataset is saved as a tensor of shape $[N, T, 2]$, where $N$ is the number of trajectories, $T$ the number of recorded steps (including initialization), and $2$ corresponds to the slope and intercept dimensions.

\section{Data Generation for MLP Weight Trajectories}
\label{sec:dg-mlp}
We generate synthetic optimization trajectories for multilayer perceptrons (MLPs) trained on a simple linear regression task \(y = a x + b + \epsilon\), where \(\epsilon \sim \mathcal{N}(0, 0.1)\). Coefficients \(a\) and \(b\) are sampled from \(\mathcal{N}(2.0, 0.01)\) and \(\mathcal{N}(1.0, 0.01)\), respectively. Each dataset consists of 100 points with \(x \in [-1, 1]\).

Two MLP architectures with exactly 15 parameters are considered: a 3-layer MLP with hidden sizes \([2, 2, 1]\) and a 2-layer MLP with hidden sizes \([4, 1]\). All models are trained using the 5 optimizers with a learning rate of 0.01 (except for adagrad with 0.1), minimizing mean squared error (MSE) over 199 epochs. Mini-batch size is set to 64. For each training run, we record the full parameter vector after every epoch, resulting in a trajectory of shape \((200, 15)\) per run (199 updates + initialization).

For reproducibility, we generate 50 trajectories per seed across 5 random seeds (0-4), where the first 30 trajectories use the 3-layer MLP and the remaining 20 use the 2-layer MLP. All experiments are conducted on an Apple MacBook Pro with an M3 Max chip (48 GB RAM) using PyTorch with MPS acceleration. This setup ensures consistent evaluation of weight evolution under varying architectures and optimizers for trajectory forecasting tasks.

\section{Neural Network Generation}\label{sec:appendix-nn-generation}

To evaluate parameter prediction across model types, we construct three separate datasets for CNNs  and Transformers:
\begin{itemize}
    \item CNNs follow the NAS-Bench201 paradigm \cite{dong2020bench}. Each architecture is defined by a 4-node directed acyclic graph (cell), where each edge chooses from 5 operations: \texttt{none}, \texttt{skip-connect}, \texttt{1x1 conv}, \texttt{3x3 conv}, or \texttt{3x3 avg-pool}. Nodes aggregate incoming features by summation. We sample 100 unique such cells.

    \item Transformers reuse the MLP sampling grid (depth, width, dropout), but each layer contains a residual attention block with:
    Multi-head Self-Attention $\rightarrow$ LayerNorm $\rightarrow$ MLP.  We fix the number of heads = 2 and set MLP expansion ratio = 2×. The  image dataset is flattened and projected to a token vector using a linear embedding. We sample depth, number of hidden neurons and dropout rate from the following ranges.
    \begin{itemize}
        \item Depth: number of hidden layers $ \in\{1, 2, 3, 4\}$
        \item Hidden size:  $\in \{64, 128, 256, 512, 1024\}$
        \item Dropout: $\in\{0.0, 0.1, 0.3\}$
    \end{itemize}     
  All layers in a given MLP share the same hidden dimension and dropout rate.
\end{itemize}     

 For each architecture type, we train 100 instances initialized with different random seeds. We log parameter values at every epoch, along with train/validation loss and accuracy. All training is performed on CIFAR-10 with consistent optimizer and learning rate schedules. The number of parameters sampled for each architecture family is 5,418 and 924,554 respectively.

For this neural network  task, the weight prediction model is trained on a single NVIDIA A40 Graphics Card.

\section{Parameter Sensitivity}
\label{sec:param-sen}
To investigate the impact of hyperparameters on model performance, we conducted a sensitivity analysis of our model on synthetic linear regression tasks. Specifically, we varied three key regularization parameters: $\beta$ , $\gamma$ , and $\zeta$, and examined their influence on prediction accuracy.

The sensitivity analysis was performed over a wide range of values: $\beta \in \{0.0, 0.1, 1.0, 10.0\}$, $\gamma \in \{0.0, 0.1, 1.0, 10.0\}$ and $\zeta \in \{0.0, 1.0, 10.0, 100.0\}$.

For each combination of hyperparameters and optimizer, we trained the model for 1000 epochs with a fixed learning rate of 0.0001, and evaluated the residual forecasting error on the held-out test set. Each experiment was repeated with five random seeds to account for variance due to initialization.

\paragraph{Results.} The results reveal the following key trends:
\begin{itemize}
    \item Small non-zero values of $\beta$ and $\gamma$ (e.g., $0.1$ or $1.0$) generally improve model generalization by encouraging smoother temporal dynamics and better alignment with ground-truth trajectories.

    \item The gradient penalty $\zeta$ exhibits beneficial effects up to moderate values (e.g., $\zeta = 1.0$ or $10.0$), consistently improving generalization across optimizers. However, excessively large values (e.g., $\zeta = 100.0$) may lead to diminishing returns or even degrade performance, particularly for SGD and Adam, due to over-regularization effects.

    \item Sensitivity varies across optimizers. For instance, AdamW benefits more from $\gamma$ regularization, while SGD shows consistent improvements with moderate $\zeta$.

\end{itemize}

Overall, the optimal configurations are found at moderate regularization strengths rather than extreme values. A detailed breakdown of the best-performing settings for each optimizer is provided in Table~\ref{tab:sensitivity-best}. A list of various configurations are shown in Table \ref{tab:zeta=1}, \ref{tab:zeta=10} and \ref{tab:zeta=100}.

\begin{table}[ht]
\centering
\caption{Best configurations for each optimizer}
\begin{tabular}{lccccc}
\toprule
Optimizer & Best Mean (Std) & Beta & Gamma & Zeta \\
\midrule
SGD       & 0.015 (0.006) & 0.0 & 0.0 & 10.0  \\
Adam      & 0.014 (0.004) & 0.0 & 0.1 & 1.0   \\
AdamW     & 0.014 (0.005) & 0.0 & 1.0 & 100.0 \\
RMSProp   & 0.015 (0.003) & 0.1 & 1.0 & 100.0 \\
Adagrad   & 0.013 (0.002) & 0.1 & 1.0 & 100.0 \\
\bottomrule
\end{tabular}
\label{tab:sensitivity-best}
\end{table}

\begin{table}[ht]
\centering
\small
\begin{tabular}{cc p{2.5cm} p{3cm}}
\toprule
$\beta$ & $\gamma$ & Optimizer & Mean (Std) \\
\midrule
\multirow{5}{*}{0.1} & \multirow{5}{*}{0.1} 
    & SGD     & 0.040 (0.034) \\
    &         & Adam    & 0.045 (0.027) \\
    &         & AdamW   & 0.031 (0.015) \\
    &         & RMSProp & 0.026 (0.021) \\
    &         & Adagrad & 0.017 (0.004) \\
\midrule
\multirow{5}{*}{0.1} & \multirow{5}{*}{1.0} 
    & SGD     & 0.033 (0.007) \\
    &         & Adam    & 0.039 (0.016) \\
    &         & AdamW   & 0.035 (0.010) \\
    &         & RMSProp & 0.025 (0.006) \\
    &         & Adagrad & 0.022 (0.007) \\
\midrule
\multirow{5}{*}{0.1} & \multirow{5}{*}{10.0} 
    & SGD     & 0.047 (0.030) \\
    &         & Adam    & 0.050 (0.022) \\
    &         & AdamW   & 0.051 (0.032) \\
    &         & RMSProp & 0.033 (0.016) \\
    &         & Adagrad & 0.030 (0.013) \\
\midrule
\multirow{5}{*}{1.0} & \multirow{5}{*}{0.1} 
    & SGD     & 0.041 (0.023) \\
    &         & Adam    & 0.051 (0.026) \\
    &         & AdamW   & 0.050 (0.023) \\
    &         & RMRProp & 0.031 (0.021) \\
    &         & Adagrad & 0.023 (0.012) \\
\midrule
\multirow{5}{*}{1.0} & \multirow{5}{*}{1.0} 
    & SGD     & 0.123 (0.035) \\
    &         & Adam    & 0.152 (0.031) \\
    &         & AdamW   & 0.193 (0.119) \\
    &         & RMSProp & 0.071 (0.006) \\
    &         & Adagrad & 0.045 (0.013) \\
\midrule
\multirow{5}{*}{1.0} & \multirow{5}{*}{10.0} 
    & SGD     & 0.144 (0.033) \\
    &         & Adam    & 0.199 (0.025) \\
    &         & AdamW   & 0.161 (0.044) \\
    &         & RMSProp & 0.074 (0.020) \\
    &         & Adagrad & 0.046 (0.013) \\
\midrule
\multirow{5}{*}{10.0} & \multirow{5}{*}{0.1} 
    & SGD     & 0.057 (0.032) \\
    &         & Adam    & 0.083 (0.040) \\
    &         & AdamW   & 0.080 (0.039) \\
    &         & RMSProp & 0.100 (0.073) \\
    &         & Adagrad & 0.087 (0.059) \\
\midrule
\multirow{5}{*}{10.0} & \multirow{5}{*}{1.0} 
    & SGD     & 0.131 (0.038) \\
    &         & Adam    & 0.210 (0.048) \\
    &         & AdamW   & 0.206 (0.041) \\
    &         & RMSProp & 0.097 (0.061) \\
    &         & Adagrad & 0.073 (0.035) \\
\midrule
\multirow{5}{*}{10.0} & \multirow{5}{*}{10.0} 
    & SGD     & 0.529 (0.122) \\
    &         & Adam    & 0.713 (0.133) \\
    &         & AdamW   & 0.724 (0.201) \\
    &         & RMSProp & 0.308 (0.106) \\
    &         & Adagrad & 0.213 (0.086) \\
\bottomrule
\end{tabular}
\caption{Performance metrics for $\zeta = 1.0$}
\label{tab:zeta=1}
\end{table}

\begin{table}[ht]
\centering
\small
\begin{tabular}{cc l l}
\toprule
$\beta$ & $\gamma$ & Optimizer & Mean (Std) \\
\midrule
\multirow{5}{*}{0.1} & \multirow{5}{*}{0.1} 
    & SGD     & 0.018 (0.010) \\
    &         & Adam    & 0.021 (0.013) \\
    &         & AdamW   & 0.023 (0.021) \\
    &         & RMSProp & 0.020 (0.013) \\
    &         & Adagrad & 0.018 (0.012) \\
\midrule
\multirow{5}{*}{0.1} & \multirow{5}{*}{1.0} 
    & SGD     & 0.017 (0.005) \\
    &         & Adam    & 0.020 (0.009) \\
    &         & AdamW   & 0.018 (0.004) \\
    &         & RMSProp & 0.017 (0.004) \\
    &         & Adagrad & 0.014 (0.004) \\
\midrule
\multirow{5}{*}{0.1} & \multirow{5}{*}{10.0} 
    & SGD     & 0.023 (0.010) \\
    &         & Adam    & 0.022 (0.008) \\
    &         & AdamW   & 0.019 (0.005) \\
    &         & RMSProp & 0.019 (0.005) \\
    &         & Adagrad & 0.019 (0.006) \\
\midrule
\multirow{5}{*}{1.0} & \multirow{5}{*}{0.1} 
    & SGD     & 0.028 (0.020) \\
    &         & Adam    & 0.029 (0.016) \\
    &         & AdamW   & 0.039 (0.043) \\
    &         & RMSProp & 0.024 (0.021) \\
    &         & Adagrad & 0.026 (0.028) \\
\midrule
\multirow{5}{*}{1.0} & \multirow{5}{*}{1.0} 
    & SGD     & 0.033 (0.019) \\
    &         & Adam    & 0.050 (0.034) \\
    &         & AdamW   & 0.033 (0.017) \\
    &         & RMSProp & 0.025 (0.017) \\
    &         & Adagrad & 0.019 (0.009) \\
\midrule
\multirow{5}{*}{1.0} & \multirow{5}{*}{10.0} 
    & SGD     & 0.035 (0.008) \\
    &         & Adam    & 0.038 (0.017) \\
    &         & AdamW   & 0.034 (0.008) \\
    &         & RMSProp & 0.025 (0.006) \\
    &         & Adagrad & 0.021 (0.008) \\
\midrule
\multirow{5}{*}{10.0} & \multirow{5}{*}{0.1} 
    & SGD     & 0.031 (0.023) \\
    &         & Adam    & 0.044 (0.048) \\
    &         & AdamW   & 0.040 (0.034) \\
    &         & RMSProp & 0.027 (0.021) \\
    &         & Adagrad & 0.025 (0.020) \\
\midrule
\multirow{5}{*}{10.0} & \multirow{5}{*}{1.0} 
    & SGD     & 0.051 (0.036) \\
    &         & Adam    & 0.048 (0.015) \\
    &         & AdamW   & 0.052 (0.031) \\
    &         & RMSProp & 0.032 (0.026) \\
    &         & Adagrad & 0.021 (0.010) \\
\midrule
\multirow{5}{*}{10.0} & \multirow{5}{*}{10.0} 
    & SGD     & 0.131 (0.018) \\
    &         & Adam    & 0.168 (0.051) \\
    &         & AdamW   & 0.150 (0.018) \\
    &         & RMSProp & 0.070 (0.021) \\
    &         & Adagrad & 0.043 (0.008) \\
\bottomrule
\end{tabular}
\caption{Performance metrics for $\zeta = 10.0$}
\label{tab:zeta=10}
\end{table}

\begin{table}[htbp]
\centering
\small
\begin{tabular}{cc l l}
\toprule
$\beta$ & $\gamma$ & Optimizer & Mean (Std) \\
\midrule
\multirow{5}{*}{0.1} & \multirow{5}{*}{0.1}
    & SGD     & 0.026 (0.015) \\
    &         & Adam    & 0.027 (0.016) \\
    &         & AdamW   & 0.021 (0.016) \\
    &         & RMSProp & 0.020 (0.011) \\
    &         & Adagrad & 0.015 (0.005) \\
\midrule
\multirow{5}{*}{0.1} & \multirow{5}{*}{1.0}
    & SGD     & 0.017 (0.004) \\
    &         & Adam    & 0.021 (0.011) \\
    &         & AdamW   & 0.015 (0.006) \\
    &         & RMSProp & 0.015 (0.003) \\
    &         & Adagrad & 0.013 (0.002) \\
\midrule
\multirow{5}{*}{0.1} & \multirow{5}{*}{10.0}
    & SGD     & 0.018 (0.008) \\
    &         & Adam    & 0.017 (0.005) \\
    &         & AdamW   & 0.015 (0.004) \\
    &         & RMSProp & 0.016 (0.005) \\
    &         & Adagrad & 0.014 (0.004) \\
\midrule
\multirow{5}{*}{1.0} & \multirow{5}{*}{0.1}
    & SGD     & 0.019 (0.008) \\
    &         & Adam    & 0.021 (0.011) \\
    &         & AdamW   & 0.014 (0.003) \\
    &         & RMSProp & 0.017 (0.005) \\
    &         & Adagrad & 0.018 (0.010) \\
\midrule
\multirow{5}{*}{1.0} & \multirow{5}{*}{1.0}
    & SGD     & 0.021 (0.013) \\
    &         & Adam    & 0.020 (0.012) \\
    &         & AdamW   & 0.016 (0.007) \\
    &         & RMSProp & 0.019 (0.011) \\
    &         & Adagrad & 0.013 (0.002) \\
\midrule
\multirow{5}{*}{1.0} & \multirow{5}{*}{10.0}
    & SGD     & 0.018 (0.008) \\
    &         & Adam    & 0.018 (0.005) \\
    &         & AdamW   & 0.020 (0.007) \\
    &         & RMSProp & 0.017 (0.004) \\
    &         & Adagrad & 0.014 (0.003) \\
\midrule
\multirow{5}{*}{10.0} & \multirow{5}{*}{0.1}
    & SGD     & 0.025 (0.017) \\
    &         & Adam    & 0.028 (0.019) \\
    &         & AdamW   & 0.032 (0.028) \\
    &         & RMSProp & 0.025 (0.017) \\
    &         & Adagrad & 0.023 (0.022) \\
\midrule
\multirow{5}{*}{10.0} & \multirow{5}{*}{1.0}
    & SGD     & 0.029 (0.024) \\
    &         & Adam    & 0.034 (0.027) \\
    &         & Adamw   & 0.027 (0.013) \\
    &         & RMSProp & 0.029 (0.030) \\
    &         & Adagrad & 0.019 (0.012) \\
\midrule
\multirow{5}{*}{10.0} & \multirow{5}{*}{10.0}
    & SGD     & 0.047 (0.051) \\
    &         & Adam    & 0.039 (0.029) \\
    &         & AdamW   & 0.045 (0.027) \\
    &         & RMSProp & 0.034 (0.031) \\
    &         & Adagrad & 0.020 (0.010) \\
\bottomrule
\end{tabular}
\caption{Performance metrics for $\zeta = 100.0$}
\label{tab:zeta=100}
\end{table}

\end{document}